\documentclass[letterpaper]{article} 
\usepackage{aaai2026}  
\usepackage{times}  
\usepackage{helvet}  
\usepackage{courier}  
\usepackage[hyphens]{url}  
\usepackage{graphicx} 
\usepackage{natbib}  
\usepackage{caption} 
\usepackage{algorithm}
\usepackage{algorithmic}
\usepackage{amsmath, amssymb}
\usepackage{subfigure}
\usepackage{booktabs}

\usepackage{newfloat}
\usepackage{listings}
\DeclareCaptionStyle{ruled}{labelfont=normalfont,labelsep=colon,strut=off} 
\floatstyle{ruled}
\newfloat{listing}{tb}{lst}{}
\floatname{listing}{Listing}
\title{Distillation Dynamics: Towards Understanding Feature-Based Distillation \\ in Vision Transformers}
\author {
	Huiyuan Tian\textsuperscript{\rm 1},
	Bonan Xu\textsuperscript{\rm 2},
	Shijian Li\textsuperscript{\rm 1}\thanks{Corresponding author} 
}
\affiliations {
	\textsuperscript{\rm 1}College of Computer Science and Technology, Zhejiang University, Hangzhou, China\\
	\textsuperscript{\rm 2}Department of Aeronautical and Aviation Engineering, The Hong Kong Polytechnic University\\
	tianhuiyuan@zju.edu.cn, bonan.xu@polyu.edu.hk, shijianli@zju.edu.cn
}

\usepackage{bibentry}

\begin{document}

\maketitle

\begin{abstract}
	While feature-based knowledge distillation has proven highly effective for compressing CNNs, these techniques unexpectedly fail when applied to Vision Transformers (ViTs), often performing worse than simple logit-based distillation. We provide the first comprehensive analysis of this phenomenon through a novel analytical framework termed as ``distillation dynamics", combining frequency spectrum analysis, information entropy metrics, and activation magnitude tracking. Our investigation reveals that ViTs exhibit a distinctive U-shaped information processing pattern: initial compression followed by expansion. We identify the root cause of negative transfer in feature distillation: a fundamental representational paradigm mismatch between teacher and student models. Through frequency-domain analysis, we show that teacher models employ distributed, high-dimensional encoding strategies in later layers that smaller student models cannot replicate due to limited channel capacity. This mismatch causes late-layer feature alignment to actively harm student performance. Our findings reveal that successful knowledge transfer in ViTs requires moving beyond naive feature mimicry to methods that respect these fundamental representational constraints, providing essential theoretical guidance for designing effective ViTs compression strategies. All source code and experimental logs are provided at \url{https://github.com/thy960112/Distillation-Dynamics}.
\end{abstract}


\section{Introduction}
ViTs \cite{dosovitskiy2021vit, liu2021swin, han2022survey} have revolutionized computer vision, achieving state-of-the-art performance across diverse tasks \cite{caron2021emerging, li2024transformer, bar-shalom2024subgraphormer} by leveraging self-attention mechanisms \cite{vaswani2017attention} to model long-range dependencies. However, their remarkable capabilities come with significant drawbacks: prohibitive computational complexity and excessive data requirements, which severely limit their deployment on resource-constrained devices and in data-scarce scenarios.

Knowledge distillation (KD) \cite{hinton2015kd, park2019relational, Tian2020Contrastive, zhao2022decoupled, yang2023knowledge, li2024detkds, son2025maskedkd} offers a promising solution to these challenges by transferring knowledge from large, complex teacher models to smaller, more efficient student models \cite{choudhary2020comprehensive}. Originally proposed by Hinton et al. \shortcite{hinton2015kd} for compressing neural networks through logit-based distillation, this technique has evolved significantly. During the CNN era, researchers extended distillation beyond output logits to include intermediate feature representations \cite{romero2015fitnet, zagoruyko2017at, heo2019comprehensive, chen2021reviewkd, ji2021show, pham2024frequency}, capturing richer information from teacher models. Through simple convolutions or projection layers, feature-based distillation enabled substantial performance improvements in student CNNs.

Given this success, researchers naturally attempted to apply feature-based distillation methods to Vision Transformers. Surprisingly, however, a striking phenomenon emerged: the simple feature distillation techniques that worked well for CNNs actually degrade performance when applied to ViTs, often performing worse than logit-based distillation \cite{touvron2021deit, miles2024vkd, sun2024logit}. While recent work such as ViTKD \cite{yang2024vitkd} has proposed ViT-specific distillation methods and observed similar failures, the fundamental reasons behind this degradation remain poorly understood. This gap in understanding severely limits our ability to design effective distillation strategies for the ViT architectures.

The core challenge lies in the fundamental architectural and representational differences between CNNs and ViTs. While sophisticated feature distillation methods \cite{hao2022manifold, yang2024vitkd, miles2024srd, fan2024scalekd, feng2025alignkd} have shown improvements, they often rely on empirical observations rather than principled theoretical guidance \cite{cheng2024layerwise}. What is urgently needed is a deep, mechanistic understanding of how information flows through ViT architectures  and why conventional feature distillation fails—insights that can guide the development of more effective distillation methods.

In this work, we conduct a comprehensive theoretical investigation of the ViTs distillation process by introducing novel analytical tools to dissect the information dynamics within these models. We employ frequency spectrum analysis, Shannon entropy analysis through the lens of the Information Bottleneck principle \cite{tishby2000ib, tishby2015dib, hong2025cib}, and activation magnitude analysis to reveal the unique information processing patterns of ViTs. Our analysis uncovers fundamental insights about how ViTs process information differently from CNNs, explaining the systematic failure of naive feature distillation approaches.

Our main contributions are as follows:
\begin{enumerate}
	\item \textbf{Novel analytical framework for understanding ViTs representations.} We introduce and systematically apply frequency spectrum analysis along the channel dimension, Shannon entropy analysis, and activation magnitude tracking to characterize the internal dynamics of ViTs. 
	\item \textbf{Discovery of the U-shaped information processing signature.} We reveal that ViT models exhibit a distinctive U-shaped pattern in information entropy across layers, indicating a two-phase processing strategy: initial compression followed by task-specific expansion. This pattern emerges during training and represents a learned, rather than architectural, behavior.
	\item \textbf{Identification of channel-dimension encoding saturation.} We discover that ViTs fully utilize their channel dimensions for distributed encoding in later layers, revealing that performance degradation in student networks stems not from insufficient parameters, but from fundamental representational capacity constraints that prevent smaller models from replicating the teacher's encoding strategy.
	\item \textbf{Mechanistic explanation of negative transfer.} Through analysis of distillation evolution, we demonstrate that late-layer feature distillation induces negative transfer by forcing students to adopt representational paradigms incompatible with their limited capacity. We show that information processing patterns evolve from monotonic to U-shaped during training, and that misaligned distillation disrupts this natural progression.
	
\end{enumerate}

These insights provide the theoretical foundation necessary for designing principled feature distillation methods for Vision Transformers, moving beyond empirical trial-and-error toward theory-guided approaches that respect the fundamental information processing characteristics of these architectures.

\section{Analytical Methods} \label{sec:md}

To dissect the internal mechanisms of knowledge distillation in ViTs, we devise a multi-faceted analytical framework termed as ``distillation dynamics". This framework is designed to characterize how a student model learns to mimic the internal representations of a teacher model by examining three complementary aspects of their activation patterns. First, we propose frequency spectrum analysis to reveal the composition of feature encodings, distinguishing between global, coarse-grained structures and fine-grained, local details. Second, we apply Shannon entropy \cite{ash2012information} analysis to quantify the information complexity and structural organization within feature maps. Finally, we measure activation magnitudes to further prove our findings by tracking the strength of signal propagation through the network layers.

These methods allow us to triangulate the true nature of ViTs representations: when we observe a U-shaped entropy pattern indicating compression followed by expansion, we can verify this through corresponding changes in frequency spectra (from uniform to low-pass to uniform again) and activation magnitudes (decreasing then increasing). This multi-faceted validation is crucial for establishing that the observed patterns represent fundamental computational strategies rather than artifacts of any single measurement approach.

\subsection{Frequency Spectrum Analysis} \label{sec:spectrum}

To understand how the network encodes features, we analyze the frequency content of the feature representations at each layer. Such analysis is particularly relevant for ViTs, which have been characterized as behaving like low-pass filters \cite{park2022how}, a property that can lead to the loss of high-frequency information in deeper layers.

Intermediate activations are extracted from ViT model during forward inference on validation images. Let the activations be denoted as a tensor tensor $\mathbf{A} \in \mathbb{R}^{L \times B \times C \times H \times W}$, where $L$ is the number of layers, $B$ is the batch size, $C$ is the number of channels, and $H \times W$ is the spatial resolution.

To reveal the complexity of feature interactions within the channel dimension, we apply a one-dimensional Fast Fourier Transform (FFT) along the channel axis for each spatial position. This unconventional choice, distinct from a spatial FFT, allows us to assess the structure of the feature space itself. A low-frequency-dominant spectrum suggests that channels are highly correlated, representing smooth variations of a primary feature, whereas a high-frequency spectrum indicates a more complex, decorrelated set of feature detectors. The transform is defined as:
\begin{equation}
	\mathbf{F}_{l,b,h,w}[k] = \frac{1}{C} \sum_{c=0}^{C-1} \mathbf{A}_{l,b,c,h,w} \, e^{-j 2\pi k c / C}, 
\end{equation}
where $\mathbf{F} \in \mathbb{C}^{L \times B \times C \times H \times W}$ is the resulting frequency-domain representation, and $j = \sqrt{-1}$ is the imaginary unit. The magnitude spectrum is computed as $|\mathbf{F}|$ and averaged over the batch ($B$), height ($H$), and width ($W$) dimensions to obtain the representative per-layer frequency spectrum:
\begin{equation}
	\mathbf{S}_l[k] = \frac{1}{B H W} \sum_{b=1}^{B} \sum_{h=1}^{H} \sum_{w=1}^{W} |\mathbf{F}_{l,b,k,h,w}|.
\end{equation}

This yields a per-layer spectral signature $\mathbf{S} \in \mathbb{R}^{L \times C}$, which reveals how feature abstraction and complexity develop throughout the network by showing whether representations are dominated by low- or high-frequency components in the channel dimension.

\subsection{Information Entropy Analysis} \label{sec:entropy}

While frequency analysis reveals the composition of learned features, we employ information entropy to quantify their structural complexity. This analysis is explicitly framed through the lens of the Information Bottleneck (IB) principle \cite{tishby2000ib, tishby2015dib, hong2025cib}. In this context, low entropy signifies a compressed, highly structured representation (a tight bottleneck), whereas high entropy indicates a more uniform or expansive distribution.

The analysis is performed on the activation tensor $\mathbf{A}$. At each spatial position $(h, w)$ for a given layer $l$ and batch item $b$, the vector of channel activations $\mathbf{v}_{l,b,h,w} \in \mathbb{R}^C$ is discretized into $N_b = 100$ bins over the global activation range $[\min(\mathbf{A}), \max(\mathbf{A})]$. This is a standard method for estimating the probability mass function of continuous activations.

From this discretization, we estimate a probability mass function $p_n$ for each bin $n$:
\begin{equation}
	p_n = \frac{h_n}{C}, \quad n = 1, \dots, N_b,
\end{equation}
where $h_n$ is the count of activations falling into bin $n$. The Shannon entropy is then computed by summing over bins with non-zero probabilities:
\begin{equation}
	E_{l,b,h,w} = -\sum_{n: p_n > 0} p_n \log_2 p_n.
\end{equation}

This procedure yields an entropy map $\mathbf{E} \in \mathbb{R}^{L \times B \times H \times W}$. We then compute the average entropy for each layer $\bar{E}_l$ by averaging across the batch and spatial dimensions:
\begin{equation}
	\bar{E}_l = \frac{1}{B H W} \sum_{b=1}^{B} \sum_{h=1}^{H} \sum_{w=1}^{W} E_{l,b,h,w}.
\end{equation}

By comparing the layer-wise entropy profiles $\{\bar{E}_l\}_{l=1}^L$ of the student and teacher models, we can quantitatively assess whether the student successfully emulates the teacher's information compression and expansion strategy. This provides a powerful tool for understanding how well the core representational dynamics are transferred.

\subsection{Activation Magnitude Analysis} \label{sec:activation}

Finally, to complement our analysis of feature content (frequency) and structure (entropy), we examine the signal propagation strength throughout the network. This is accomplished by measuring the mean activation magnitude at each layer, which serves as a proxy for how the network amplifies or attenuates information as it flows through successive layers.

Using the same activation tensor $\mathbf{A}$, the mean activation magnitude for each layer $M_l$ is defined as the mean absolute value of its activation:
\begin{equation}
	M_l = \frac{1}{B C HW} \sum_{b=1}^{B} \sum_{n=1}^{C} \sum_{d=1}^{H}\sum_{d=1}^{W} |\mathbf{A}_{l,b,c,h,w}|,
\end{equation}
where the analysis spans all $L$ layers. Plotting these per-layer magnitudes $M_l$ against the layer index reveals the macroscopic information flow within the network.

\subsection{Distillation Methods for Validation}

To validate our analytical insights, we implement two feature-based distillation methods. First, SpectralKD aligns frequency spectra by applying 2D FFT to spatial dimensions of feature maps. After channel alignment via adaptive pooling when $C_s \neq C_t$, we compute the frequency alignment loss:
\begin{equation}
	\mathcal{L}_{\mathrm{Freq}} = \mathrm{MSE}(\mathcal{F}_{\mathrm{stack}}(\mathbf{A}_s), \mathcal{F}_{\mathrm{stack}}(\mathbf{A}_t)),
\end{equation}
where $\mathcal{F}_{\mathrm{stack}}$ denotes the 2D RFFT followed by stacking real and imaginary components.

Second, ProjectorKD inspired by FitNet \cite{romero2015fitnet} uses a learnable projector to match feature dimensions directly in spatial domain:
\begin{equation}
	\mathcal{L}_{\mathrm{Proj}} = \mathrm{MSE}(\mathrm{Projector}(\mathbf{A}_s), \mathbf{A}_t).
\end{equation}

Both methods combine feature losses with standard KD loss: $\mathcal{L}_{\mathrm{Total}} = \mathcal{L}_{\mathrm{KD}} + \beta \mathcal{L}_{\mathrm{Feature}}$, where $\beta$ weights the feature component. 

\begin{figure}[t]
	\centering
	\includegraphics[width=0.95\columnwidth]{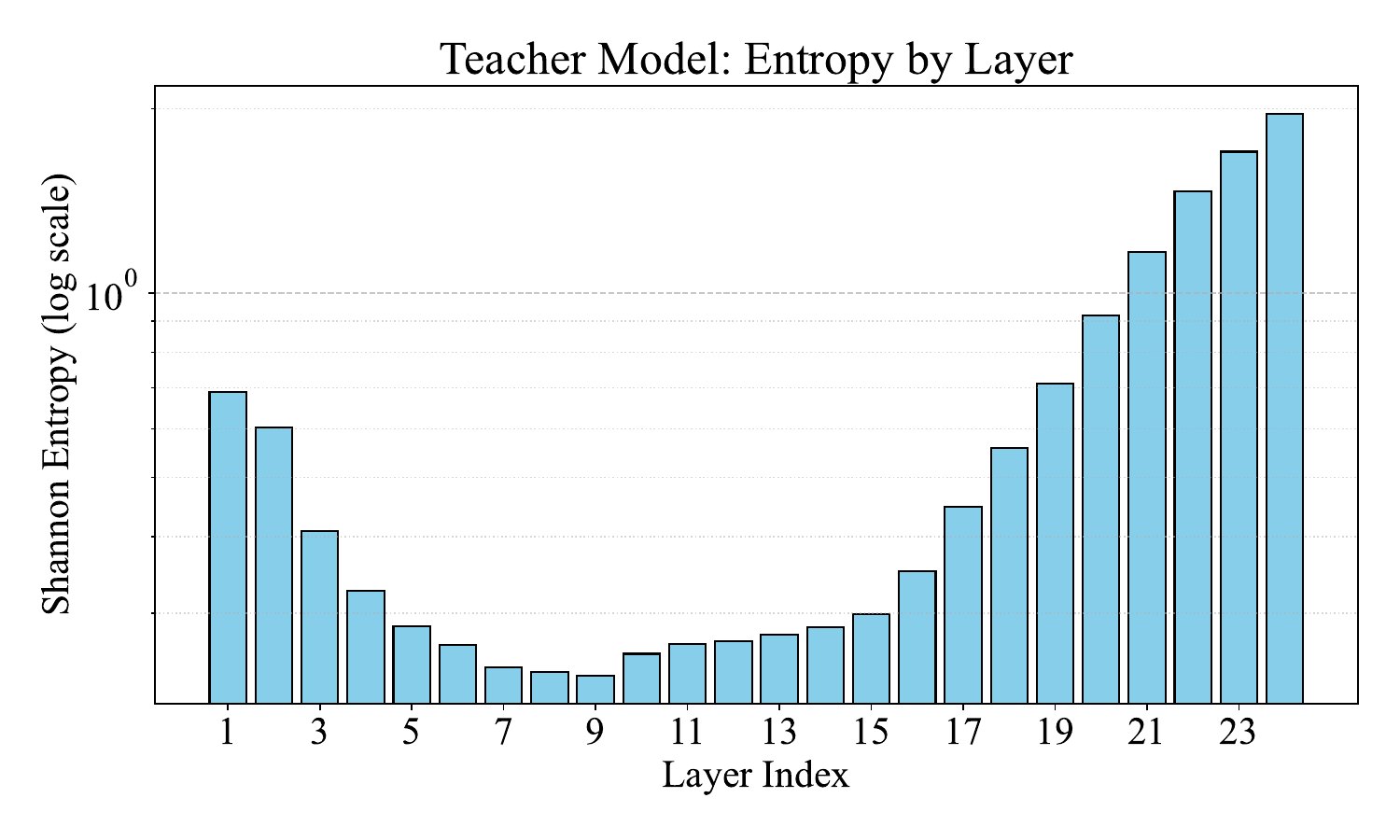} 
	\caption{Layer-wise Shannon entropy of the CaiT-S24 teacher model exhibits a characteristic U-shaped profile. Entropy decreases from layers 1-9 (compression phase), then increases through layer 24 (expansion phase).}
	\label{fig:cait}
\end{figure}

\begin{figure}[t]
	\centering
	\includegraphics[width=0.95\columnwidth]{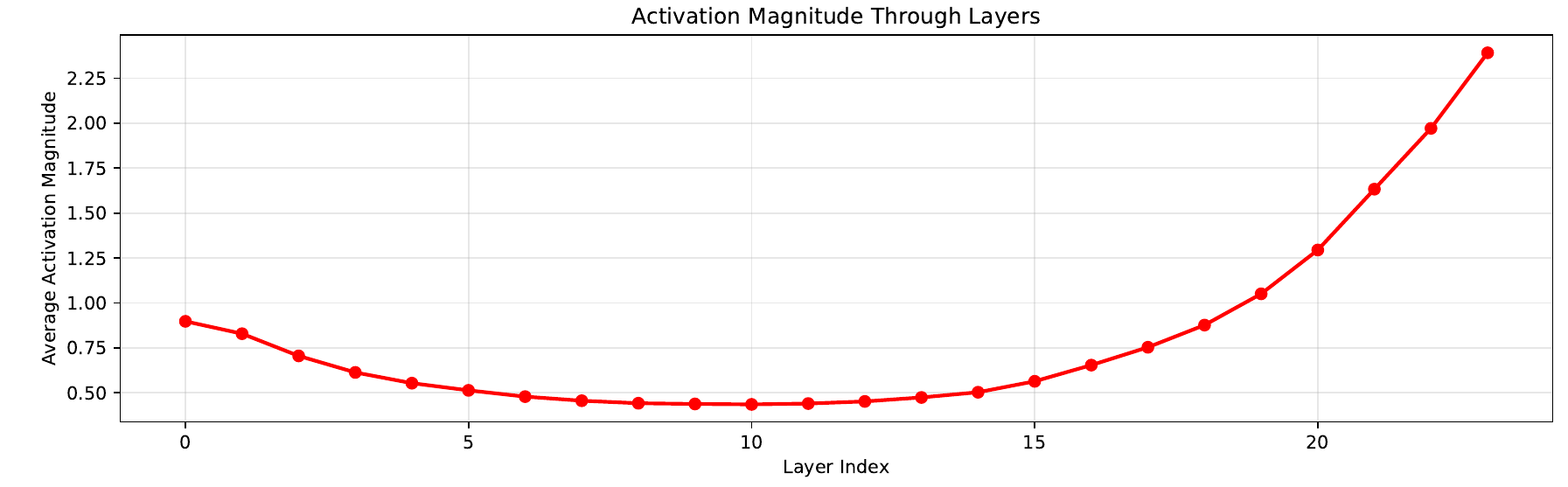} 
	\caption{U-shaped profile of average activation magnitude through network layers.}
	\label{fig:activation}
\end{figure}

\section{Analyses of Representational Dynamics} \label{sec:ana}

This section initiates our empirical investigation by first establishing a detailed characterization of the teacher model's internal information processing dynamics. By dissecting the layer-wise behavior of CaiT-S24 teacher \cite{touvron2021cait}, we define a precise information processing signature that a student model should ideally learn to replicate.

We analyze its intermediate activations generated from a forward pass on 32 randomly selected samples from the ImageNet \cite{deng2009imagenet} validation dataset. We employ the Shannon entropy and activation magnitude analyses detailed in Section~\ref{sec:md} to probe the model's layer-wise dynamics. Our findings reveal a consistent and highly structured two-phase processing pattern, which we identify as a form of representational bottleneck. The overall structure reveals an active, two-stage computational strategy: the network first builds a compressed, abstract model of the input (compression phase) and then queries this model to construct an answer for the specific task (expansion phase).

\subsection{The U-Shaped Entropy Profile}

Our analysis begins with the layer-wise Shannon entropy defined in Section~\ref{sec:entropy}. As depicted in Figure~\ref{fig:cait}, the CaiT-S24 teacher model exhibits a distinct U-shaped entropy profile across its 24 layers. The entropy initially decreases from the input layer, reaches a nadir around layers 8-9, and then steadily increases toward the final layer. Our observation of a U-shaped curve in ViTs points to a more complex, multi-stage information processing strategy that goes beyond simple, progressive feature stabilization.

This distinctive profile can be interpreted as direct empirical evidence for the IB principle operating within ViTs. The IB framework posits that an effective model learns by first compressing input data to retain only task-relevant information, and then using this compressed representation for prediction \cite{hong2025cib}. The observed U-shaped entropy curve provides clear evidence of such a two-phase strategy:
\begin{enumerate}
	\item \textbf{Compression Phase (Layers 1 to 9):} The initial, descending arm of the U-shape aligns with the compression phase of the IB. In these early layers, the model processes high-dimensional image patch embeddings. The decreasing entropy signifies that the activation distributions are becoming more concentrated, structured, and less random as the network filters out redundant visual details and noise. This process extracts and organizes essential, low-level features, which form the building blocks of the model's representation.
	\item \textbf{Refinement and Expansion Phase (Layers 9 to 24):} The subsequent, ascending arm of the curve corresponds to a refinement or expansion phase. Having formed a compact, abstract representation at the entropy nadir, the model begins to expand the feature space to construct more complex, high-level semantic concepts necessary for the final classification task. This involves combining the compressed features in myriad ways, leading to a more uniform and thus higher-entropy distribution as the model prepares the representation for the final linear classifier.
\end{enumerate}

We further present the U-shape as a consistent operational signature of the tested ViT models, as similar structures are also observable in other architectures like vanilla ViT trained by supervised learning \cite{dosovitskiy2021vit} or self-supervised learning like MAE \cite{he2022mae} (see appendix for more details). This U-shaped entropy curve marks the functional transition point where the model switches from general-purpose feature extraction to task-specific information aggregation.

\subsection{The U-Shaped Signal Propagation}

To further substantiate our finding of a two-phase model, we analyze the mean activation magnitude at each layer, as detailed in Section~\ref{sec:activation}. The resulting curve, shown in Figure~\ref{fig:activation}, strikingly mirrors the U-shaped entropy profile. The activation magnitude initially decreases, bottoms out in the middle layers, and then rises in the latter half of the network. This provides strong corroborating evidence for our findings.

This parallel U-shaped dynamic can be interpreted as follows:
\begin{enumerate}
	\item \textbf{Phase 1 (Signal Attenuation):} The initial decrease in activation magnitude suggests that during the compression phase, the model actively attenuates signals corresponding to irrelevant or redundant visual information. This functions as a form of dynamic, input-dependent feature pruning, allowing the model to focus its computational resources on more salient regions or tokens.
	\item \textbf{Phase 2 (Signal Amplification):} The subsequent increase in magnitude indicates that during the refinement phase, the model amplifies the signals of the most discriminative features. This targeted amplification ensures that the information most critical for the final decision is given the greatest weight.
\end{enumerate}

The pronounced signal amplification in the later layers can be connected to the recently identified phenomenon of massive activations \cite{sun2024massive, owen2025massive} in Transformer models. These researches have shown that a small subset of neurons in both LLMs and ViTs can exhibit exceptionally large activation magnitudes. The emergence pattern of these massive activations which appears abruptly after the initial layers and persisting until the final few layers is highly consistent with the U-shaped magnitude profile we observe.

The similar U-shaped pattern in Shannon entropy and activation magnitude together constitute the teacher's unique information processing signature. This signature reveals a sophisticated, dynamic process: the model first navigates a representational bottleneck by compressing and attenuating irrelevant information, and then expands and amplifies task-specific information for the final prediction.

\section{Results and Analyses} \label{sec:results}

This section evaluates the efficacy of transferring these internal representations to a  smaller DeiT-Tiny student model. Our experimental results reveal a series of counterintuitive phenomena, most notably the general failure of feature-based distillation to outperform a simple logits-only baseline. By dissecting these results through our proposed analytical framework, we find a fundamental representational mismatch between the teacher and student, particularly in the later layers. We further analyze the observed negative transfer in a frequency-domain, offering a new perspective on the challenges of knowledge distillation for Vision Transformers.

\begin{table}[t]
	\centering
	\begin{tabular}{llll}
		\toprule
		Method & Layer(s) & $\beta$  & Top-1 Acc (\%) \\ \midrule
		SoftKD & N/A & N/A & 76.99 \\ 
		SoftKD & N/A & N/A & 78.07 (500 epochs) \\ \hline
		SpectralKD & First 2, Last 6 & 0.2 & 77.07 \\ 
		SpectralKD & First 2, Last 1 & 0.2 & 77.06 \\ 
		SpectralKD & First 1, Last 1 & 0.2 & 77.08 \\ 
		SpectralKD & First 1 & 0.2 & 77.00 \\ 
		SpectralKD & Last 1 & 0.2 & 76.83 \\ 
		SpectralKD & Last 1 & 0.2 & 77.59 (500 epochs) \\
		SpectralKD & Last 1 & 0.1 & 76.48 \\ 
		SpectralKD & Last 8 & 0.2 & 76.69 \\ \hline
		ProjectorKD & First 1 & 0.2 & 76.86 \\ 
		ProjectorKD & Last 1 & 0.2 & 76.72 \\ 
		ProjectorKD & First 1, Last 1 & 0.2 & 76.80 \\ \bottomrule
	\end{tabular}
	\caption{Top-1 Accuracy on ImageNet for a DeiT-Tiny student distilled from a CaiT-S24 teacher. The SoftKD baseline uses only logits-based distillation. All feature-based methods (SpectralKD, ProjectorKD) are combined with logits-based distillation. ``First X" and ``Last Y" refer to aligning the first X and last Y layers of the student with corresponding layers from the teacher. $\beta$ is the weight of feature-based distillation loss.  All models are trained with 300 epochs except those noted with 500 epochs.}
	\label{tab:acc}
\end{table}

\subsection{Empirical Evaluation of Feature Distillation}

To systematically investigate the transfer of intermediate representations, we conduct a series of distillation experiments using CaiT-S24 model as the teacher and a smaller DeiT-Tiny model as the student. We evaluate three primary distillation schemes: a baseline using only logit-based distillation (SoftKD), our proposed non-parametric frequency alignment method (SpectralKD), and a parametric projector-based feature distillation method (ProjectorKD). For the feature-based methods, we explore distilling knowledge from various layers or combinations of layers. The ImageNet Top-1 accuracy are presented in Table~\ref{tab:acc}.

A puzzle observation in Table~\ref{tab:acc} is that, contrary to the prevailing assumption that intermediate features provide a richer and more informative supervisory signal, the majority of our feature distillation experiments fail to surpass the performance  of the simple SoftKD baseline ($76.99\%$) with distillation temperature set to $1$. The only configurations that yield a marginal improvement are those that incorporate knowledge from the teacher's early layers. The best model ($77.08\%$) uses SpectralKD to align the first and last layers of teacher and student models.

This outcome is consistent across both the non-parametric SpectralKD and the parametric ProjectorKD. Specifically, attempts to distill knowledge from the teacher's final layers prove to be actively detrimental. For instance, using ProjectorKD on the final layer results in a Top-1 accuracy of $76.72\%$, a drop of $0.27\%$ compared to the SoftKD baseline. Similarly, SpectralKD applied to the last layer yields $76.83\%$, and the last eight layers results in $76.69\%$. This consistent performance drop across different methods points toward a systemic issue with transferring knowledge from the teacher's late-stage representations. The phenomenon has also been observed by previous works like FitNet \cite{romero2015fitnet} and ViTKD \cite{yang2024vitkd}.

These observations strengthen our hypothesis: since two methodologically distinct approaches point to the same conclusion, the problem likely lies not in the mechanism of transfer, but in the intrinsic nature of the knowledge being transferred from different stages of the teacher model.

\subsection{Negative Transfer from Late Layers} \label{sec:paradox}

The consistent performance degradation observed when distilling from the teacher's later layers reveals a phenomenon of negative transfer. Our experiments underscore the severity of this issue through a surprising result. Initially, we hypothesize that negative transfer might occur because the feature distillation loss overpowers the primary classification loss. To test this, we reduce the feature-distillation weight $\beta$ for SpectralKD on the last layer from $0.2$ to $0.1$. Conventional wisdom would predict that this weaker, less restrictive guidance should alleviate the negative effects. However, we observe the opposite: performance drops even further, from $76.83\%$ to $76.48\%$  (Table~\ref{tab:acc}).

We also explore whether extend training may improve feature map distillation performance, reasoning that feature maps contain rich information requiring more time to learn effectively. After $500$ training epochs, simple SoftKD achieved $78.07\%$ accuracy while SpectralKD with last-layer distillation reached only $77.59\%$, widening the performance gap despite the longer training period.

This counterintuitive outcome strongly suggests that the problem with late-layer distillation is not one of magnitude but of direction. The guidance from the teacher's final layers fundamentally misdirects the student's learning trajectory, providing a supervisory signal that conflicts with the student's own optimization path.

\begin{figure}[t]
	\centering
	\begin{minipage}[b]{\linewidth}
		\subfigure[Layer $1$.]{
			\includegraphics[width=0.295\linewidth]{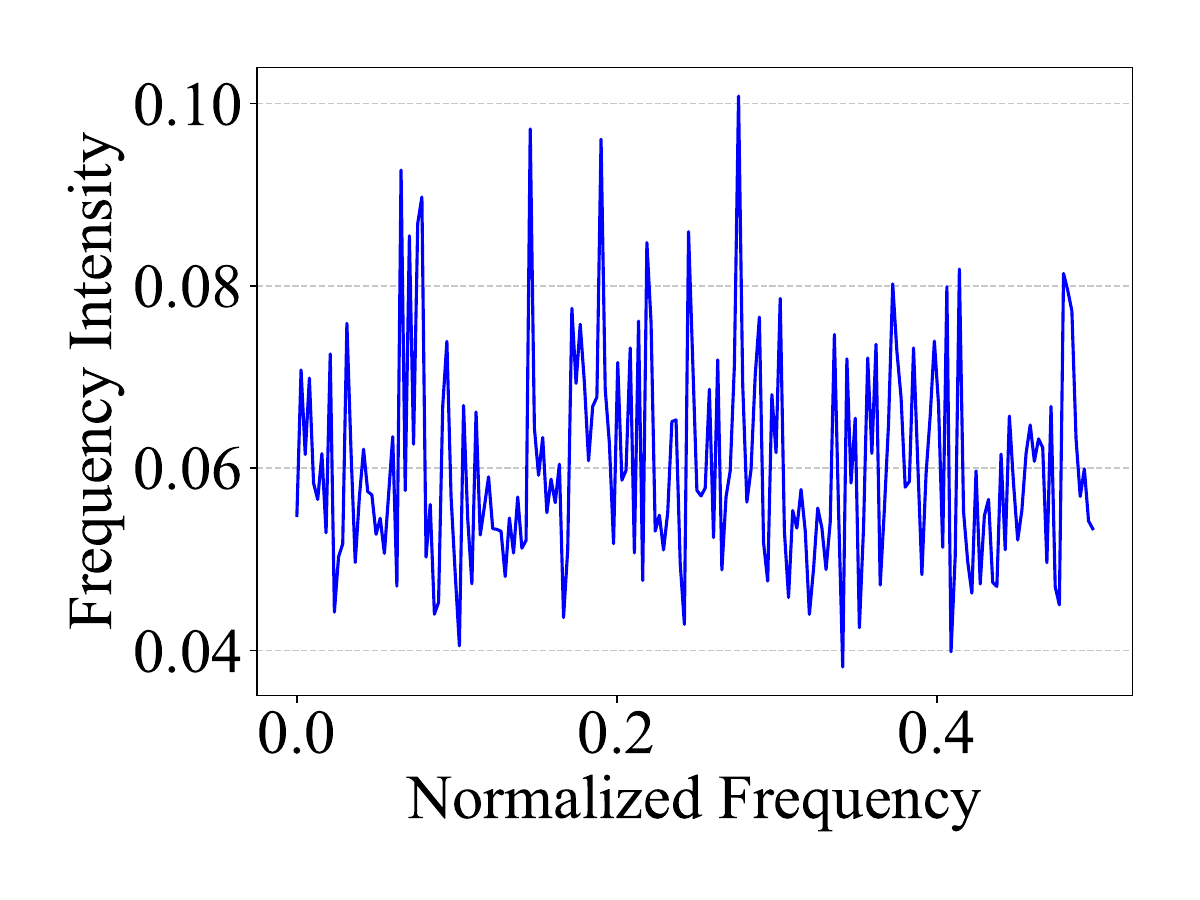}
			\label{Layer_1}
		}
		\subfigure[Layer $2$.]{
			\includegraphics[width=0.295\linewidth]{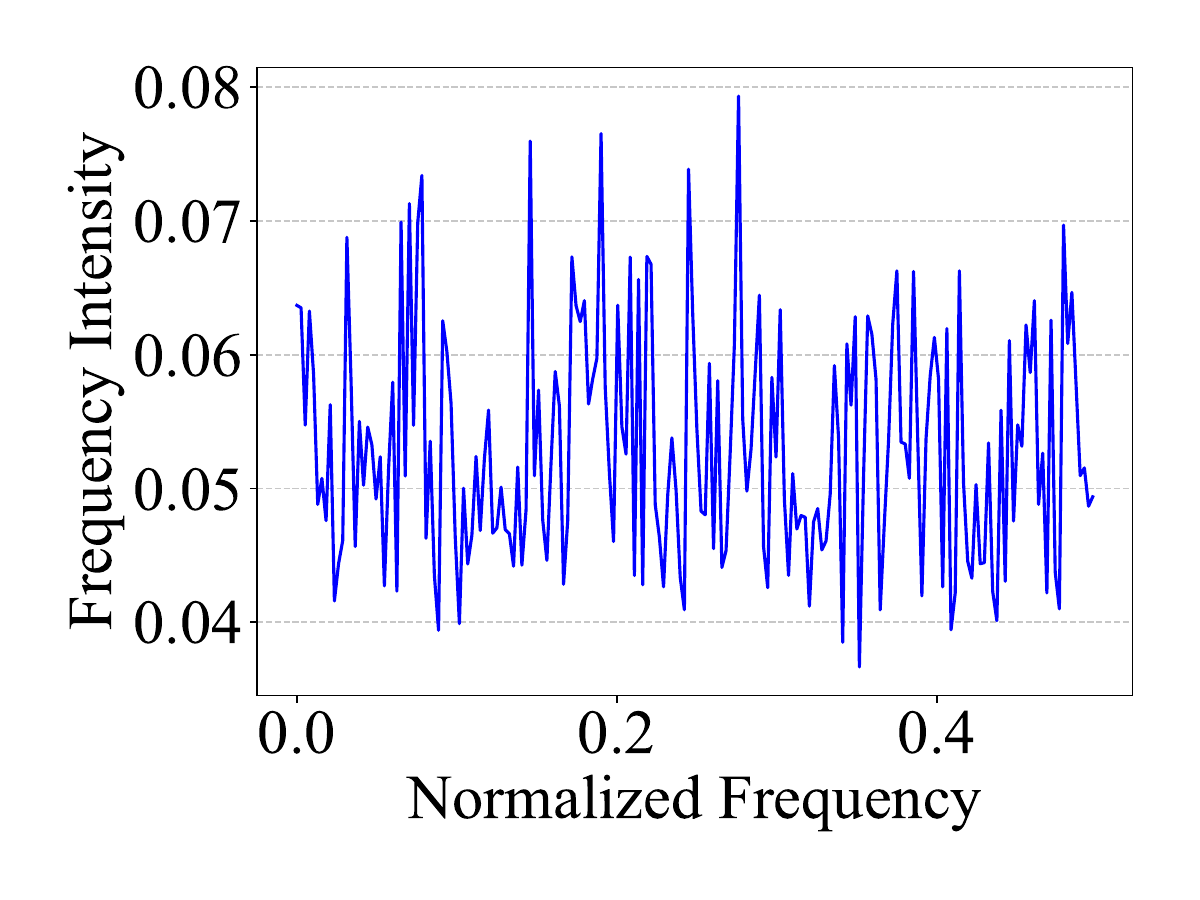}
			\label{Layer_2}
		}
		\subfigure[Layer $12$.]{
			\includegraphics[width=0.295\linewidth]{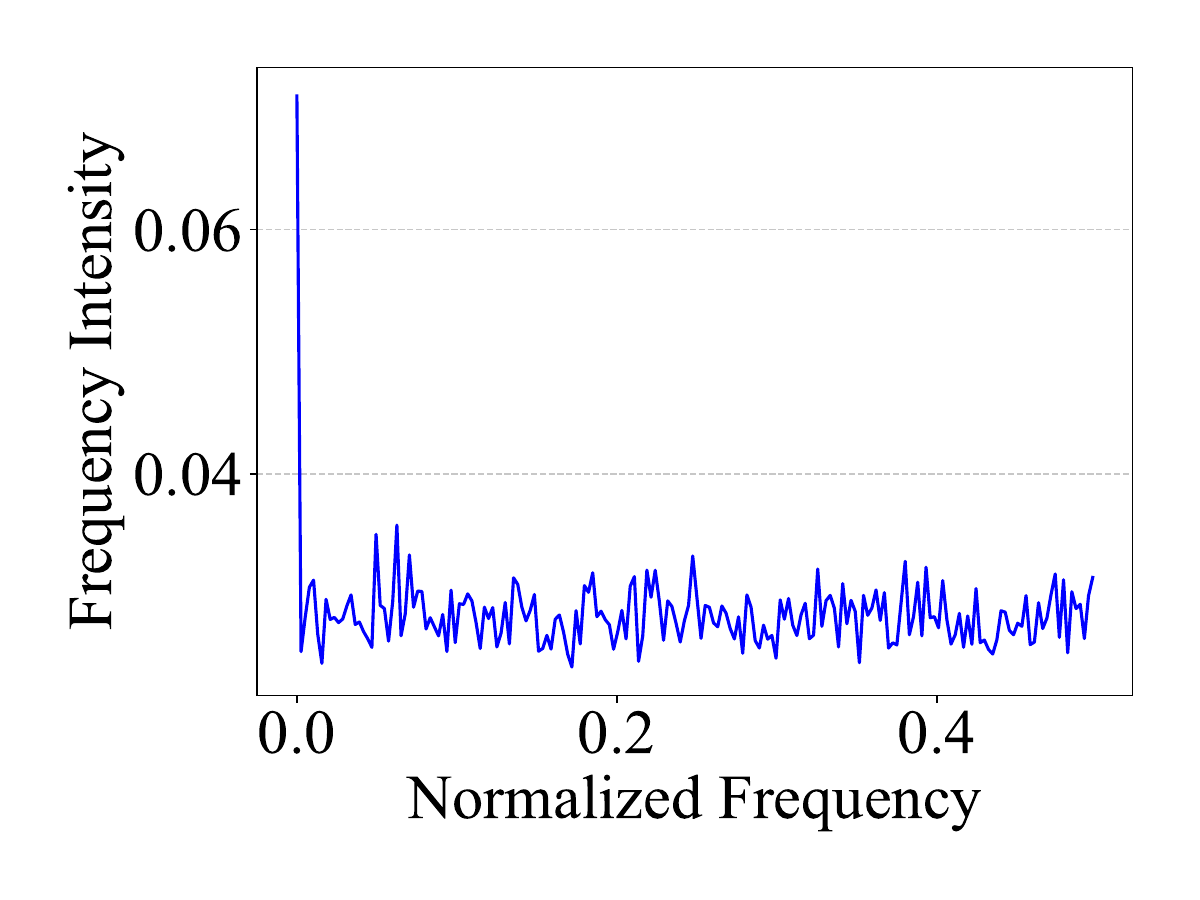}
			\label{Layer_12}
		}
	\end{minipage}
	\begin{minipage}[b]{\linewidth}
		\subfigure[Layer $13$.]{
			\includegraphics[width=0.295\linewidth]{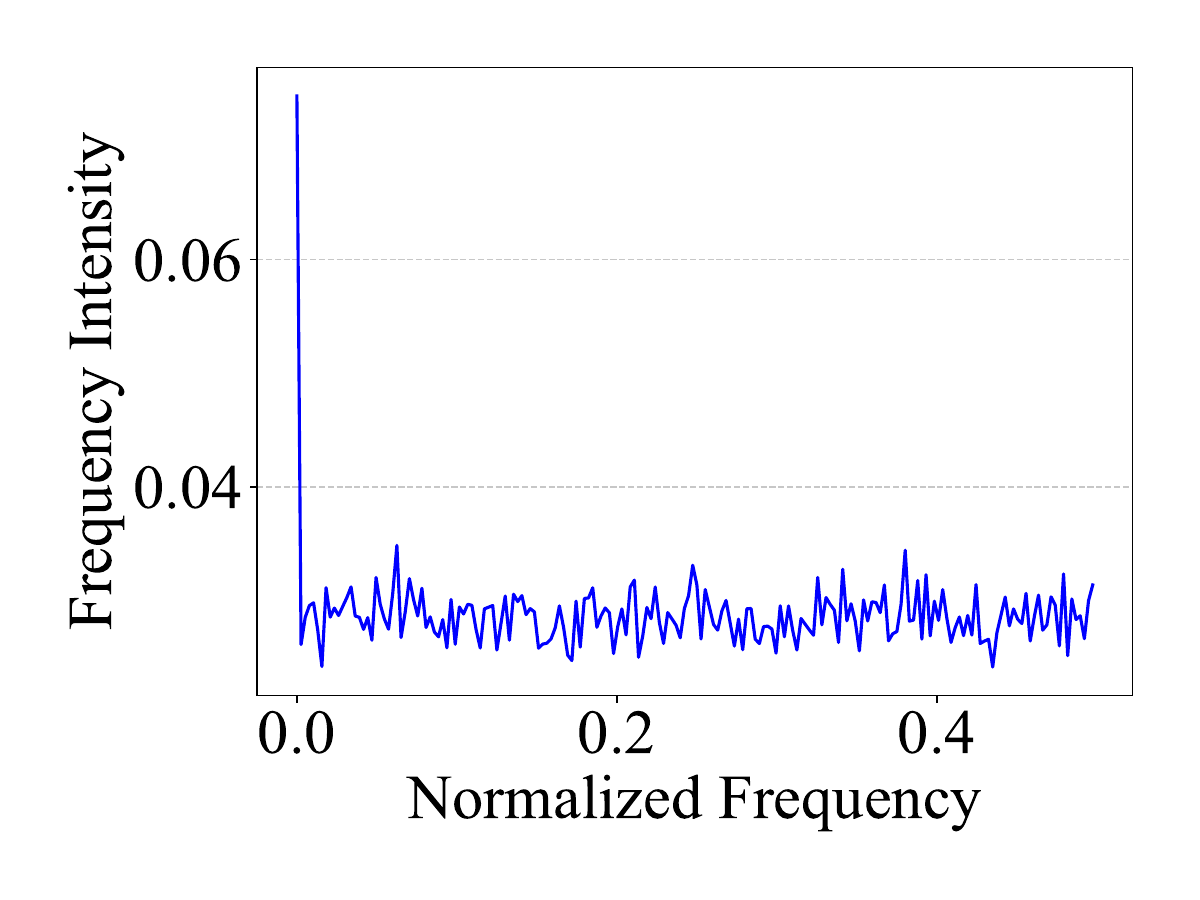}
			\label{Layer_13}
		}
		\subfigure[Layer $23$.]{
			\includegraphics[width=0.295\linewidth]{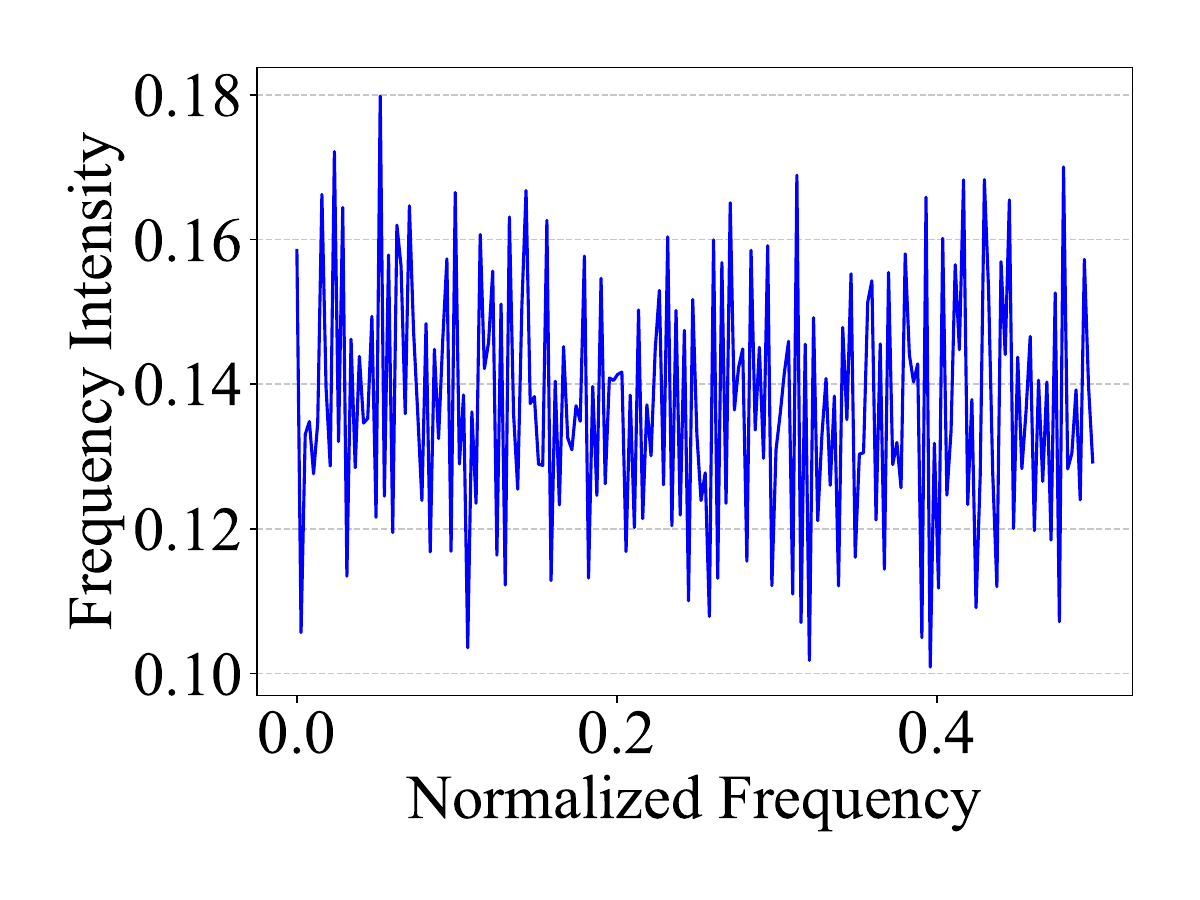}
			\label{Layer_23}
		}
		\subfigure[Layer $24$.]{
			\includegraphics[width=0.295\linewidth]{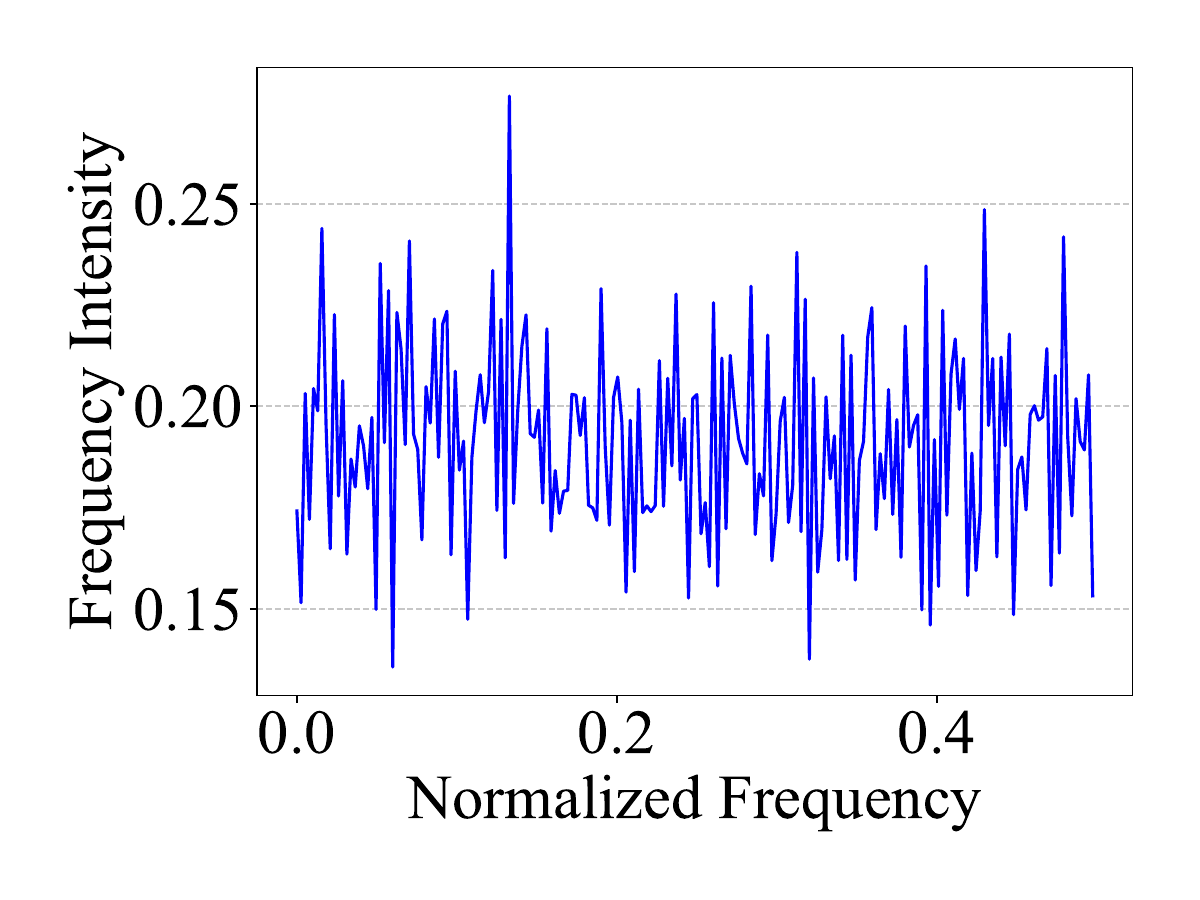}
			\label{Layer_24}
		}
	\end{minipage}
	\caption{Frequency spectra of channel-wise feature representations across layers in CaiT-S24. Early layers (a-b) show uniform, noisy spectra; middle layers (c-d) exhibit low-pass filter characteristics corresponding to the representational bottleneck; late layers (e-f) return to uniform spectra with higher energy, indicating distributed high-dimensional encoding. This three-phase evolution aligns with the U-shaped information processing profile in ViTs.}
	\label{fig:pattern}
\end{figure}

\subsection{Representational Mismatch in Frequency Domain} 

To uncover the mechanistic origin of this pathological guidance, we employ the frequency spectrum analysis detailed in Section~\ref{sec:spectrum}. By examining the spectral content of the teacher model's activations along the channel dimension at different layers, we can reveal the encoded patterns of the teacher model in the frequency domain. Figure~\ref{fig:pattern} presents the frequency spectra for representative layers of the CaiT-S24 teacher model.

The spectral signatures reveal a distinct three-phase evolution that aligns perfectly with the U-shaped information processing profile identified in Section~\ref{sec:ana}:

\begin{enumerate}
	\item \textbf{Phase 1: Early Layers (e.g., Layers 1-2).} In the initial layers, the frequency spectra are relatively uniform and noisy, with no dominant pattern. This corresponds to the high-entropy, high-magnitude beginning of the U-shaped curves. At this stage, the model processes raw patch embeddings, and the representations remain generic, containing a broad mixture of low- and high-frequency information across channels.
	\item \textbf{Phase 2: Middle Layers (e.g., Layers 12-13).} Around the middle of the network, the spectra exhibit a pronounced decay from low-frequency to high-frequency components, resembling a low-pass filter response. This spectral pattern coincides with the nadir of the U-shaped entropy curve, further confirming the representational bottleneck in ViTs. Here, the model has compressed the input by filtering out high-frequency noise while retaining structured, abstract representations of the most salient information.
	\item \textbf{Phase 3: Late Layers (e.g., Layers 23-24).} In the final layers, the spectrum becomes more uniform again, but with higher overall energy compared to early layers. This corresponds to the ``expansion phase" of the U-shaped entropy curve.
\end{enumerate}

The spectral pattern in Phase 3 provides critical evidence for explaining negative transfer. The return to a uniform, high-energy spectrum does not represent a regression to the noisy state of early layers. Instead, it signifies that the teacher model is performing complex, high-dimensional feature expansion, distributing and entangling semantic information across the entire channel space in an intricate manner. This sophisticated computational strategy depends intrinsically on the teacher's high representational capacity and its massive channel dimension. 

Interestingly, CNN architectures like ResNet \cite{he2016deep} exhibit different spectral patterns. Their final stages maintain spectral encoding characteristics similar to Phase 2 (middle layers) rather than the distributed encoding of Phase 3. This suggests that CNNs underutilize their channel dimension capacity, which explains why smaller student models can successfully learn CNN teacher features through distillation. Further details on CNN spectral characteristics are provided in the appendix.

This analysis reframes the problem of late-layer distillation in ViTs. The issue extends beyond a simple quantitative ``capacity gap" in parameter count or layer width. It represents a fundamental representational paradigm mismatch. The teacher's late layers operate under a distributed, high-dimensional encoding paradigm, while the student model, with its severely limited channel dimension, is architecturally incapable of replicating this approach. Instead, the student is forced to operate within a more ``compact, feature-centric" paradigm, where information must be encoded efficiently within its constrained channel space. Consequently, this knowledge becomes effectively non-transferable to smaller students through direct mimicry.

\begin{figure}[t]
	\centering
	\begin{minipage}[b]{\linewidth}
		\subfigure[SoftKD.]{
			\includegraphics[width=0.465\linewidth]{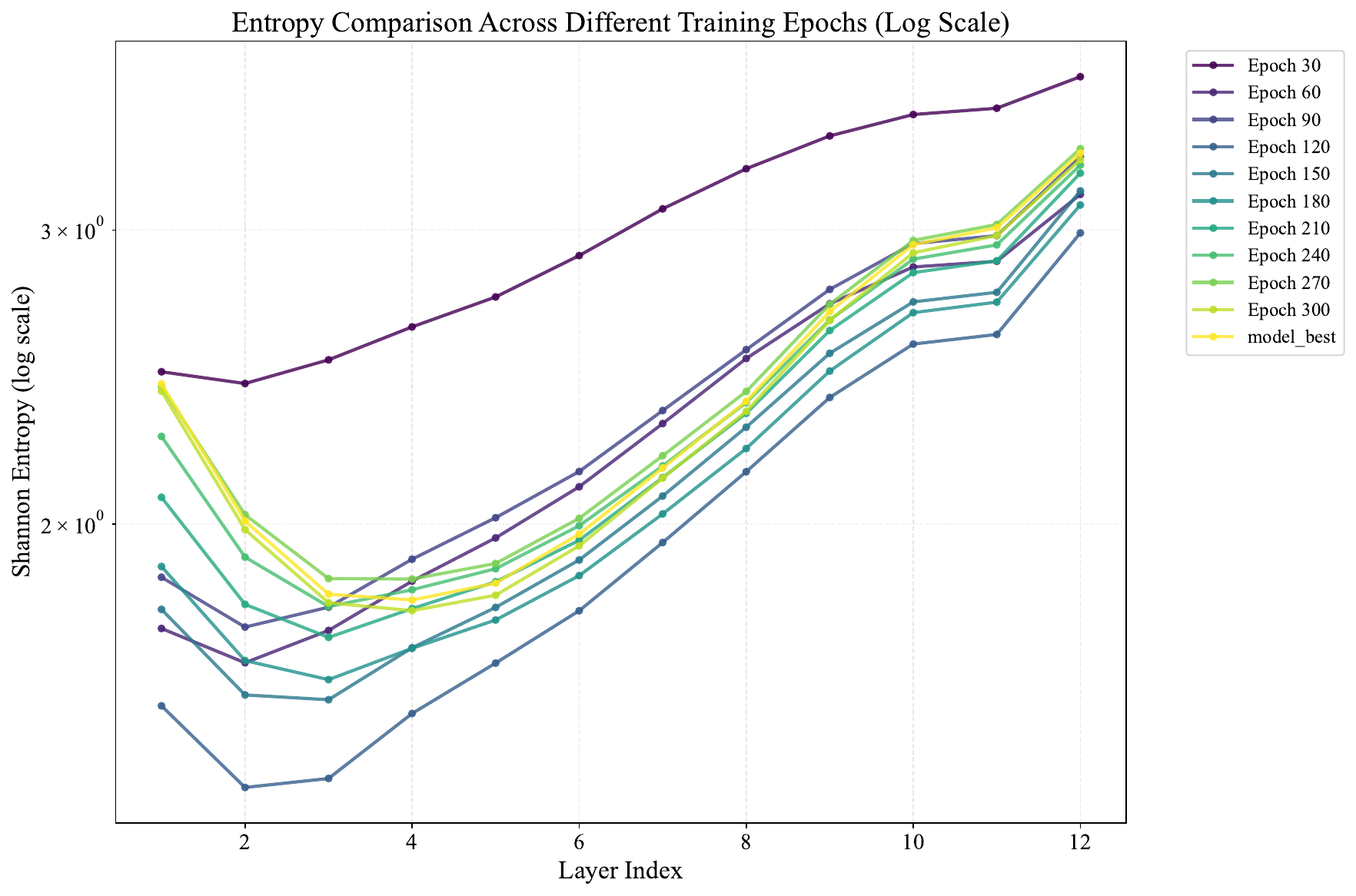}
			\label{fig:SoftKD}
		}
		\subfigure[SpectralKD-First.]{
			\includegraphics[width=0.465\linewidth]{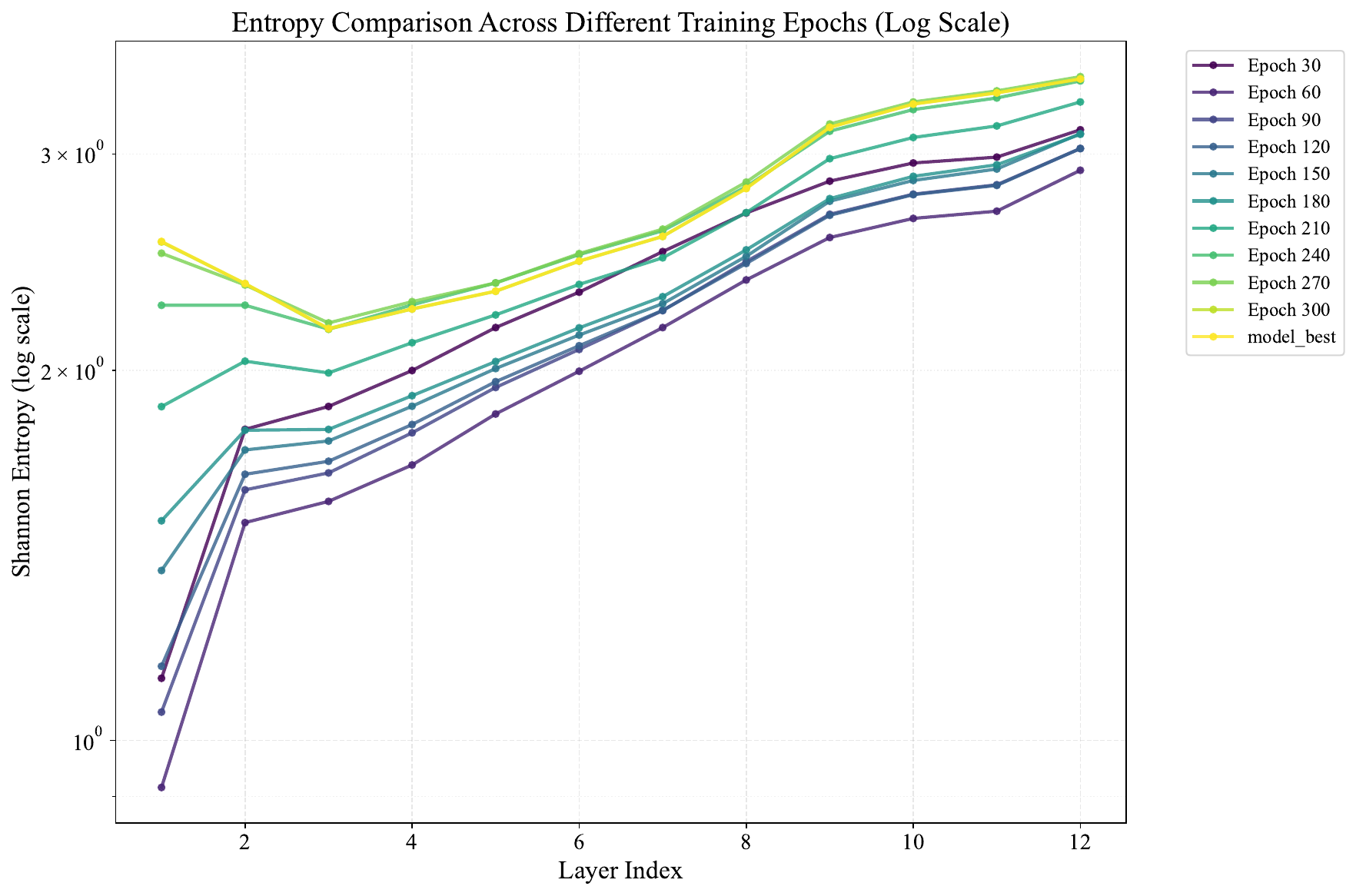}
			\label{fig:SpectralKD10}
		}
	\end{minipage}
	\begin{minipage}[b]{\linewidth}
		\subfigure[SpectralKD-Last.]{
			\includegraphics[width=0.465\linewidth]{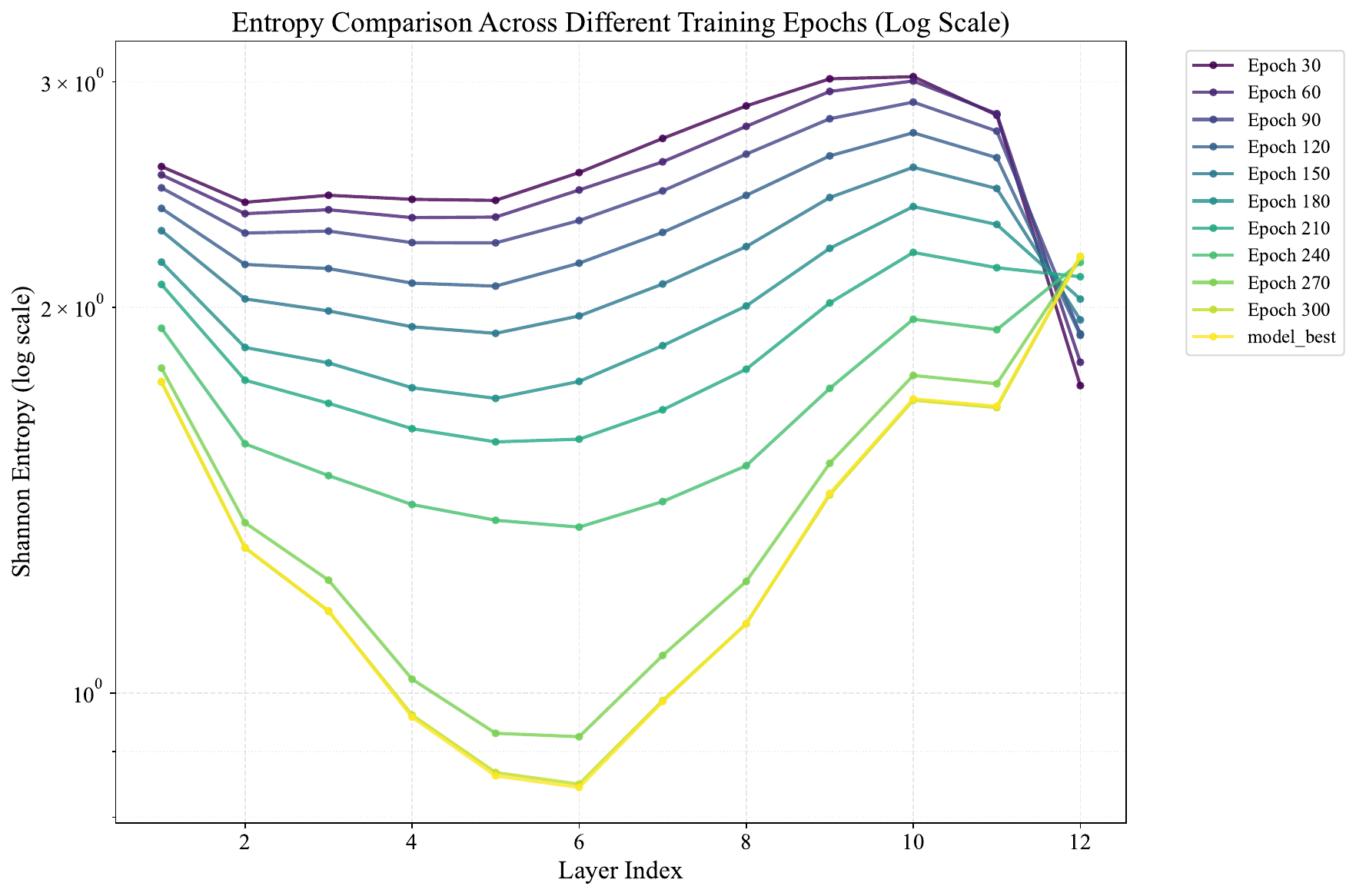}
			\label{fig:SpectralKD01}
		}
		\subfigure[SpectralKD-Both.]{
			\includegraphics[width=0.465\linewidth]{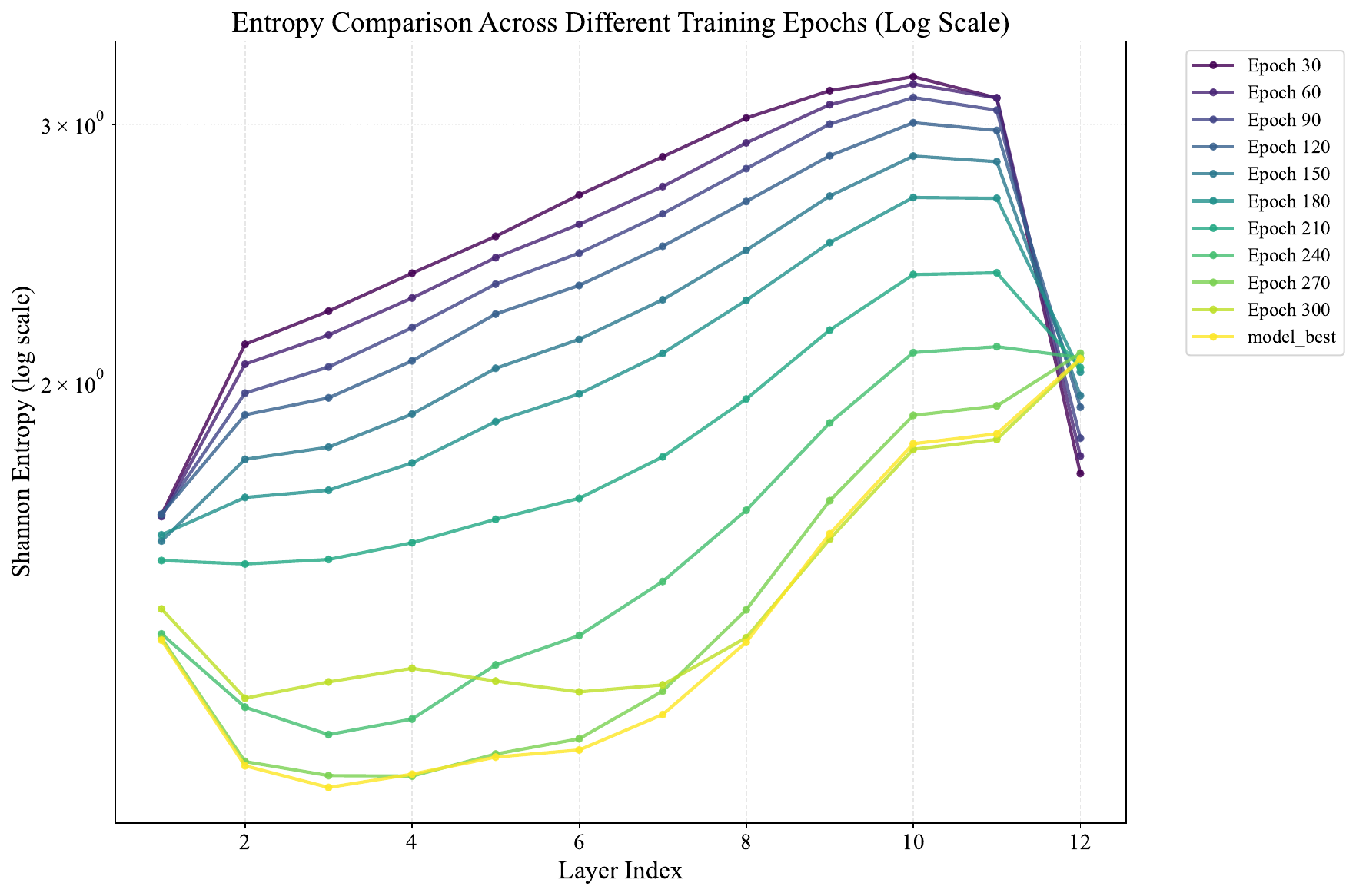}
			\label{fig:SpectralKD11}
		}
	\end{minipage}
	\caption{Distillation Evolution.}
	\label{fig:dynamics}
\end{figure}

\section{Distillation Evolution}

To deepen our understanding of the representational mismatch identified in Section~\ref{sec:results}, we examine the temporal evolution of the student's layer-wise Shannon entropy during training. This analysis reveals how distillation strategies shape the development of the student's information processing signature over epochs. By comparing the entropy profiles at checkpoints every 30 epochs (up to 300), we uncover how feature alignment alters the emergence of the U-shaped structure, explaining the observed negative transfer from late layers.

We focus on four configurations: logits-only distillation (SoftKD), frequency alignment on the first layer (SpectralKD-First), on the last layer (SpectralKD-Last), and on both the first and last layers (SpectralKD-Both). Entropy values were computed as described in Section~\ref{sec:entropy} on DeiT-Tiny students distilled from the CaiT-S24 teacher. 

\subsection{Dynamics under Logits-Only Distillation}

Under SoftKD (Figure~\ref{fig:SoftKD}), the student's entropy profile in early training (e.g., epochs 30 to 60) exhibits a predominantly increasing trend across layers. As training advances, however, the profile transforms. By epoch 120, entropy in the middle layers (e.g., layers 3 to 6) begins to dip, and by the final best checkpoint, a pronounced U-shaped structure emerges, with a minimum around layers 4 to 5. This gradual development suggests that the U-shaped signature is not an architectural artifact of ViTs but a learned behavior.

The teacher's soft labels alone, a form of response-based knowledge, suffice to guide the student toward discovering this two-phase process: initial compression followed by expansion.

\subsection{Dynamics with Late-Layer Alignment}

Incorporating frequency alignment on the last layer (SpectralKD-Last, Figure~\ref{fig:SpectralKD01}) introduces distinct perturbations to the student's developmental trajectory. In early epochs (30 to 90), the overall entropy is elevated compared to SoftKD, but the last layer starts notably lower. This initial suppression suggests that mimicking the teacher's late-layer spectral structure imposes a compressive constraint, forcing the student to adopt a more structured (lower-entropy) representation prematurely in what should be its expansion phase.

As training progresses, the last-layer entropy gradually increases, while early-layer entropy decreases. This upward drift in the last layer implies a conflict: the feature alignment loss pulls the student's representation toward the teacher's distributed, high-dimensional encoding, but the student's limited capacity and the primary classification objective push for higher entropy to enable flexible feature recombination. This dynamic highlights negative transfer as a tug-of-war, where late-layer constraints hinder the student's natural progression toward the teacher's full signature.

\subsection{Early-Layer Alignment and Global Effects}

Aligning the first layer (SpectralKD-First, Figure~\ref{fig:SpectralKD10}) yields a different trajectory. Early training shows lower entropy in initial layers, rising steadily to layer 12, which is a monotonic increase without a clear dip. Over epochs, entropy in early layers rises substantially, but the profile remains largely increasing even in the best model, lacking the teacher's U-shaped bottleneck. This suggests that early-layer alignment enforces a more uniform, high-frequency-dominant structure from the outset (aligning with the teacher's noisy Phase 1 spectra), accelerating feature extraction but preventing the compression phase from fully developing. The absence of a mid-network minimum may indicate over-specialization to low-level details, yet performance remains comparable to SoftKD ($77.00\%$), implying that early guidance complements logits without disrupting later expansion.

Combining first- and last-layer alignment (Figure~\ref{fig:SpectralKD11}) reveals compounded effects. The profile starts with moderate entropy and evolves to a shallow increase, with minimal U-shape. Strikingly, introducing late-layer alignment causes early-layer entropy to decrease over training. This global propagation, where last-layer constraints retroactively suppress early-layer entropy, demonstrates that feature distillation has global impacts. It contradicts the expected U-shape's high initial entropy, suggesting that dual constraints over-regularize both phases, flattening the bottleneck. Despite this, the marginal performance gain ($77.08\%$) indicates a delicate balance where complementary alignments may mitigate some representational mismatches.

\subsection{Implications and Insights}

These dynamics underscore that feature distillation is not a static process of copying knowledge but a dynamic process of guiding the student's developmental trajectory. It fundamentally reshapes representational development, acting as a form of curriculum learning where a mismatched curriculum can hinder performance.

A new observation is the ``entropy rebound" in late layers under last-layer alignment: the student's attempt to increase entropy despite constraints points to an adaptive resilience, where the classification loss counters over-compression from the distillation loss. This could explain why reducing the distillation weight $\beta$ worsens performance (Section~\ref{sec:paradox}). A weaker alignment allows a partial rebound toward the student's preferred high-entropy state but ultimately disrupts the learned equilibrium, leading to a poorer solution.

Furthermore, the global effects of local alignment suggest the presence of emergent, highly coupled inter-layer dependencies in ViTs, where spectral constraints can propagate bidirectionally through the network during training.

\section{Conclusion}

This work provides a mechanistic explanation for the systemic failure of feature-based distillation in ViTs. Proposing a novel analytical framework termed as ``distillation dynamics" that combines frequency, entropy, and activation analysis, we discover that ViTs learn a characteristic ``U-shaped" information processing signature, defined by an initial compression phase followed by a task-specific expansion phase. The failure of feature-based distillation stems from a fundamental representational paradigm mismatch between teacher and student models. In their later layers, large teacher models employ a sophisticated, distributed encoding strategy that entangles information across their entire high-dimensional channel space. Smaller student models, with their limited channel capacity, are architecturally incapable of replicating this strategy.

Forcing a student to mimic this late-layer representation provides a conflicting and actively harmful supervisory signal. Our analysis of the distillation process over time shows that this misguided mimicry disrupts the student's natural learning trajectory, preventing it from developing its own effective U-shaped processing pattern. In contrast, aligning only the earlier, compressive-phase layers proves more compatible and less detrimental.

These findings demonstrate that effective ViTs compression requires moving beyond naive feature mimicry. Promising directions include phase-specific distillation, which applies feature alignment only to the compatible early-to-middle layers of the teacher, and transformation-based approaches that explicitly translate the teacher's complex representations into a format the student can digest. This work provides the theoretical foundation for designing principled and effective compression strategies tailored to the unique information dynamics of ViTs.

\section{Acknowledgments}
This work was supported by the STI 2030 Major Projects (grant 2021ZD0200403) and by the Zhejiang Provincial Natural Science Foundation of China (grant LD24F030002).

\bibliography{aaai2026}

@InProceedings{cheng2024layerwise,
	title = 	 {Layerwise Change of Knowledge in Neural Networks},
	author =       {Cheng, Xu and Cheng, Lei and Peng, Zhaoran and Xu, Yang and Han, Tian and Zhang, Quanshi},
	booktitle = 	 {Proceedings of the 41st International Conference on Machine Learning},
	pages = 	 {8038--8059},
	year = 	 {2024},
	editor = 	 {Salakhutdinov, Ruslan and Kolter, Zico and Heller, Katherine and Weller, Adrian and Oliver, Nuria and Scarlett, Jonathan and Berkenkamp, Felix},
	volume = 	 {235},
	series = 	 {Proceedings of Machine Learning Research},
	month = 	 {21--27 Jul},
	publisher =    {PMLR},
	url = 	 {https://proceedings.mlr.press/v235/cheng24b.html}
}

@inproceedings{park2022how,
title={How Do Vision Transformers Work?},
author={Namuk Park and Songkuk Kim},
booktitle={International Conference on Learning Representations},
year={2022},
url={https://openreview.net/forum?id=D78Go4hVcxO}
}

@misc{hinton2015kd,
      title={Distilling the Knowledge in a Neural Network}, 
      author={Geoffrey Hinton and Oriol Vinyals and Jeff Dean},
      year={2015},
      eprint={1503.02531},
      archivePrefix={arXiv},
      primaryClass={stat.ML},
      url={https://arxiv.org/abs/1503.02531}, 
}

@misc{romero2015fitnet,
      title={FitNets: Hints for Thin Deep Nets}, 
      author={Adriana Romero and Nicolas Ballas and Samira Ebrahimi Kahou and Antoine Chassang and Carlo Gatta and Yoshua Bengio},
      year={2015},
      eprint={1412.6550},
      archivePrefix={arXiv},
      primaryClass={cs.LG},
      url={https://arxiv.org/abs/1412.6550}, 
}

@inproceedings{zagoruyko2017at,
title={Paying More Attention to Attention: Improving the Performance of Convolutional Neural Networks via Attention Transfer},
author={Sergey Zagoruyko and Nikos Komodakis},
booktitle={International Conference on Learning Representations},
year={2017},
url={https://openreview.net/forum?id=Sks9_ajex}
}

@inproceedings{chen2021reviewkd,
    author    = {Chen, Pengguang and Liu, Shu and Zhao, Hengshuang and Jia, Jiaya},
    title     = {Distilling Knowledge via Knowledge Review},
    booktitle = {Proceedings of the IEEE/CVF Conference on Computer Vision and Pattern Recognition (CVPR)},
    month     = {June},
    year      = {2021},
    pages     = {5008-5017}
}

@article{ji2021show, 
title={Show, Attend and Distill: Knowledge Distillation via Attention-based Feature Matching}, volume={35}, url={https://ojs.aaai.org/index.php/AAAI/article/view/16969}, DOI={10.1609/aaai.v35i9.16969}, abstractNote={Knowledge distillation extracts general knowledge from a pretrained teacher network and provides guidance to a target student network. Most studies manually tie intermediate features of the teacher and student, and transfer knowledge through predefined links. However, manual selection often constructs ineffective links that limit the improvement from the distillation. There has been an attempt to address the problem, but it is still challenging to identify effective links under practical scenarios. In this paper, we introduce an effective and efficient feature distillation method utilizing all the feature levels of the teacher without manually selecting the links. Specifically, our method utilizes an attention based meta network that learns relative similarities between features, and applies identified similarities to control distillation intensities of all possible pairs. As a result, our method determines competent links more efficiently than the previous approach and provides better performance on model compression and transfer learning tasks. Further qualitative analyses and ablative studies describe how our method contributes to better distillation.}, number={9}, journal={Proceedings of the AAAI Conference on Artificial Intelligence}, author={Ji, Mingi and Heo, Byeongho and Park, Sungrae}, year={2021}, month={May}, pages={7945-7952} 
}

@inproceedings{hao2022manifold,
 author = {Hao, Zhiwei and Guo, Jianyuan and Jia, Ding and Han, Kai and Tang, Yehui and Zhang, Chao and Hu, Han and Wang, Yunhe},
 booktitle = {Advances in Neural Information Processing Systems},
 editor = {S. Koyejo and S. Mohamed and A. Agarwal and D. Belgrave and K. Cho and A. Oh},
 pages = {9164--9175},
 publisher = {Curran Associates, Inc.},
 title = {Learning Efficient Vision Transformers via Fine-Grained Manifold Distillation},
 url = {https://proceedings.neurips.cc/paper_files/paper/2022/file/3bd2d73b4e96b0ac5a319be58a96016c-Paper-Conference.pdf},
 volume = {35},
 year = {2022}
}

@inproceedings{yang2024vitkd,
    author    = {Yang, Zhendong and Li, Zhe and Zeng, Ailing and Li, Zexian and Yuan, Chun and Li, Yu},
    title     = {ViTKD: Feature-based Knowledge Distillation for Vision Transformers},
    booktitle = {Proceedings of the IEEE/CVF Conference on Computer Vision and Pattern Recognition (CVPR) Workshops},
    month     = {June},
    year      = {2024},
    pages     = {1379-1388}
}

@inproceedings{fan2024scalekd,
 author = {Fan, Jiawei and Li, Chao and Liu, Xiaolong and Yao, Anbang},
 booktitle = {Advances in Neural Information Processing Systems},
 editor = {A. Globerson and L. Mackey and D. Belgrave and A. Fan and U. Paquet and J. Tomczak and C. Zhang},
 pages = {63290--63315},
 publisher = {Curran Associates, Inc.},
 title = {ScaleKD: Strong Vision Transformers Could Be Excellent Teachers},
 url = {https://proceedings.neurips.cc/paper_files/paper/2024/file/73bda9a20f6f9f6074ce822e76f126bb-Paper-Conference.pdf},
 volume = {37},
 year = {2024}
}

@InProceedings{feng2025alignkd,
    author    = {Feng, Qianhan and Li, Wenshuo and Lin, Tong and Chen, Xinghao},
    title     = {Align-KD: Distilling Cross-Modal Alignment Knowledge for Mobile Vision-Language Large Model Enhancement},
    booktitle = {Proceedings of the IEEE/CVF Conference on Computer Vision and Pattern Recognition (CVPR)},
    month     = {June},
    year      = {2025},
    pages     = {4178-4188}
}

@article{miles2024srd, 
title={Understanding the Role of the Projector in Knowledge Distillation}, volume={38}, url={https://ojs.aaai.org/index.php/AAAI/article/view/28219}, DOI={10.1609/aaai.v38i5.28219}, abstractNote={In this paper we revisit the efficacy of knowledge distillation as a function matching and metric learning problem. In doing so we verify three important design decisions, namely the normalisation, soft maximum function, and projection layers as key ingredients. We theoretically show that the projector implicitly encodes information on past examples, enabling relational gradients for the student. We then show that the normalisation of representations is tightly coupled with the training dynamics of this projector, which can have a large impact on the students performance. Finally, we show that a simple soft maximum function can be used to address any significant capacity gap problems. Experimental results on various benchmark datasets demonstrate that using these insights can lead to superior or comparable performance to state-of-the-art knowledge distillation techniques, despite being much more computationally efficient. In particular, we obtain these results across image classification (CIFAR100 and ImageNet), object detection (COCO2017), and on more difficult distillation objectives, such as training data efficient transformers, whereby we attain a 77.2% top-1 accuracy with DeiT-Ti on ImageNet. Code and models are publicly available.}, number={5}, journal={Proceedings of the AAAI Conference on Artificial Intelligence}, author={Miles, Roy and Mikolajczyk, Krystian}, year={2024}, month={Mar.}, pages={4233-4241} 
}

@inproceedings{vaswani2017attention,
author = {Vaswani, Ashish and Shazeer, Noam and Parmar, Niki and Uszkoreit, Jakob and Jones, Llion and Gomez, Aidan N. and Kaiser, \L{}ukasz and Polosukhin, Illia},
title = {Attention is all you need},
year = {2017},
isbn = {9781510860964},
publisher = {Curran Associates Inc.},
address = {Red Hook, NY, USA},
booktitle = {Proceedings of the 31st International Conference on Neural Information Processing Systems},
pages = {6000–6010},
numpages = {11},
location = {Long Beach, California, USA},
series = {NIPS'17}
}

@inproceedings{dosovitskiy2021vit,
title={An Image is Worth 16x16 Words: Transformers for Image Recognition at Scale},
author={Alexey Dosovitskiy and Lucas Beyer and Alexander Kolesnikov and Dirk Weissenborn and Xiaohua Zhai and Thomas Unterthiner and Mostafa Dehghani and Matthias Minderer and Georg Heigold and Sylvain Gelly and Jakob Uszkoreit and Neil Houlsby},
booktitle={International Conference on Learning Representations},
year={2021},
url={https://openreview.net/forum?id=YicbFdNTTy}
}

@INPROCEEDINGS{he2022mae,
  author={He, Kaiming and Chen, Xinlei and Xie, Saining and Li, Yanghao and Dollár, Piotr and Girshick, Ross},
  booktitle={2022 IEEE/CVF Conference on Computer Vision and Pattern Recognition (CVPR)}, 
  title={Masked Autoencoders Are Scalable Vision Learners}, 
  year={2022},
  volume={},
  number={},
  pages={15979-15988},
  doi={10.1109/CVPR52688.2022.01553}
  }

@inproceedings{touvron2021deit,
  title = 	 {Training data-efficient image transformers \&amp; distillation through attention},
  author =       {Touvron, Hugo and Cord, Matthieu and Douze, Matthijs and Massa, Francisco and Sablayrolles, Alexandre and Jegou, Herve},
  booktitle = 	 {Proceedings of the 38th International Conference on Machine Learning},
  pages = 	 {10347--10357},
  year = 	 {2021},
  editor = 	 {Meila, Marina and Zhang, Tong},
  volume = 	 {139},
  series = 	 {Proceedings of Machine Learning Research},
  month = 	 {18--24 Jul},
  publisher =    {PMLR},
  url = 	 {https://proceedings.mlr.press/v139/touvron21a.html}
}

@inproceedings{touvron2021cait,
    author    = {Touvron, Hugo and Cord, Matthieu and Sablayrolles, Alexandre and Synnaeve, Gabriel and J\'egou, Herv\'e},
    title     = {Going Deeper With Image Transformers},
    booktitle = {Proceedings of the IEEE/CVF International Conference on Computer Vision (ICCV)},
    month     = {October},
    year      = {2021},
    pages     = {32-42}
}

@misc{tishby2000ib,
      title={The information bottleneck method}, 
      author={Naftali Tishby and Fernando C. Pereira and William Bialek},
      year={2000},
      eprint={physics/0004057},
      archivePrefix={arXiv},
      primaryClass={physics.data-an},
      url={https://arxiv.org/abs/physics/0004057}, 
}

@inproceedings{tishby2015dib,
  author={Tishby, Naftali and Zaslavsky, Noga},
  booktitle={2015 IEEE Information Theory Workshop (ITW)}, 
  title={Deep learning and the information bottleneck principle}, 
  year={2015},
  volume={},
  number={},
  pages={1-5},
  doi={10.1109/ITW.2015.7133169}}

@InProceedings{hong2025cib,
    author    = {Hong, Jung-Ho and Kim, Ho-Joong and Jeon, Kyu-Sung and Lee, Seong-Whan},
    title     = {Comprehensive Information Bottleneck for Unveiling Universal Attribution to Interpret Vision Transformers},
    booktitle = {Proceedings of the IEEE/CVF Conference on Computer Vision and Pattern Recognition (CVPR)},
    month     = {June},
    year      = {2025},
    pages     = {25166-25175}
}

@inproceedings{sun2024massive,
title={Massive Activations in Large Language Models},
author={Mingjie Sun and Xinlei Chen and J Zico Kolter and Zhuang Liu},
booktitle={First Conference on Language Modeling},
year={2024},
url={https://openreview.net/forum?id=F7aAhfitX6}
}

@misc{owen2025massive,
      title={A Refined Analysis of Massive Activations in LLMs}, 
      author={Louis Owen and Nilabhra Roy Chowdhury and Abhay Kumar and Fabian Güra},
      year={2025},
      eprint={2503.22329},
      archivePrefix={arXiv},
      primaryClass={cs.CL},
      url={https://arxiv.org/abs/2503.22329}, 
}

@inproceedings{pham2024frequency,
  title={Frequency attention for knowledge distillation},
  author={Pham, Cuong and Nguyen, Van-Anh and Le, Trung and Phung, Dinh and Carneiro, Gustavo and Do, Thanh-Toan},
  booktitle={Proceedings of the IEEE/CVF Winter Conference on Applications of Computer Vision},
  pages={2277--2286},
  year={2024}
}

@article{li2024transformer,
  title={Transformer-based visual segmentation: A survey},
  author={Li, Xiangtai and Ding, Henghui and Yuan, Haobo and Zhang, Wenwei and Pang, Jiangmiao and Cheng, Guangliang and Chen, Kai and Liu, Ziwei and Loy, Chen Change},
  journal={IEEE Transactions on Pattern Analysis and Machine Intelligence},
  year={2024},
  publisher={IEEE}
}

@article{han2022survey,
  title={A survey on vision transformer},
  author={Han, Kai and Wang, Yunhe and Chen, Hanting and Chen, Xinghao and Guo, Jianyuan and Liu, Zhenhua and Tang, Yehui and Xiao, An and Xu, Chunjing and Xu, Yixing and others},
  journal={IEEE transactions on pattern analysis and machine intelligence},
  volume={45},
  number={1},
  pages={87--110},
  year={2022},
  publisher={IEEE}
}

@article{choudhary2020comprehensive,
  title={A comprehensive survey on model compression and acceleration},
  author={Choudhary, Tejalal and Mishra, Vipul and Goswami, Anurag and Sarangapani, Jagannathan},
  journal={Artificial Intelligence Review},
  volume={53},
  pages={5113--5155},
  year={2020},
  publisher={Springer}
}

@inproceedings{bar-shalom2024subgraphormer,
title={Subgraphormer: Unifying Subgraph {GNN}s and Graph Transformers via Graph Products},
author={Guy Bar-Shalom and Beatrice Bevilacqua and Haggai Maron},
booktitle={Forty-first International Conference on Machine Learning},
year={2024},
}

@inproceedings{caron2021emerging,
  title={Emerging properties in self-supervised vision transformers},
  author={Caron, Mathilde and Touvron, Hugo and Misra, Ishan and J{\'e}gou, Herv{\'e} and Mairal, Julien and Bojanowski, Piotr and Joulin, Armand},
  booktitle={Proceedings of the IEEE/CVF international conference on computer vision},
  pages={9650--9660},
  year={2021}
}

@inproceedings{li2024detkds,
title={Det{KDS}: Knowledge Distillation Search for Object Detectors},
author={Lujun Li and Yufan Bao and Peijie Dong and Chuanguang Yang and Anggeng Li and Wenhan Luo and Qifeng Liu and Wei Xue and Yike Guo},
booktitle={Forty-first International Conference on Machine Learning},
year={2024},
}

@inproceedings{yang2023knowledge,
	title={From knowledge distillation to self-knowledge distillation: A unified approach with normalized loss and customized soft labels},
	author={Yang, Zhendong and Zeng, Ailing and Li, Zhe and Zhang, Tianke and Yuan, Chun and Li, Yu},
	booktitle={Proceedings of the IEEE/CVF International Conference on Computer Vision},
	pages={17185--17194},
	year={2023}
}

@book{ash2012information,
  title={Information theory},
  author={Ash, Robert B},
  year={2012},
  publisher={Courier Corporation}
}

@inproceedings{deng2009imagenet,
  title={Imagenet: A large-scale hierarchical image database},
  author={Deng, Jia and Dong, Wei and Socher, Richard and Li, Li-Jia and Li, Kai and Fei-Fei, Li},
  booktitle={2009 IEEE conference on computer vision and pattern recognition},
  pages={248--255},
  year={2009},
  organization={Ieee}
}

@inproceedings{Tian2020Contrastive,
title={Contrastive Representation Distillation},
author={Yonglong Tian and Dilip Krishnan and Phillip Isola},
booktitle={International Conference on Learning Representations},
year={2020},
url={https://openreview.net/forum?id=SkgpBJrtvS}
}

@inproceedings{zhao2022decoupled,
  title={Decoupled knowledge distillation},
  author={Zhao, Borui and Cui, Quan and Song, Renjie and Qiu, Yiyu and Liang, Jiajun},
  booktitle={Proceedings of the IEEE/CVF Conference on computer vision and pattern recognition},
  pages={11953--11962},
  year={2022}
}

@inproceedings{park2019relational,
  title={Relational knowledge distillation},
  author={Park, Wonpyo and Kim, Dongju and Lu, Yan and Cho, Minsu},
  booktitle={Proceedings of the IEEE/CVF conference on computer vision and pattern recognition},
  pages={3967--3976},
  year={2019}
}

@inproceedings{heo2019comprehensive,
  title={A comprehensive overhaul of feature distillation},
  author={Heo, Byeongho and Kim, Jeesoo and Yun, Sangdoo and Park, Hyojin and Kwak, Nojun and Choi, Jin Young},
  booktitle={Proceedings of the IEEE/CVF international conference on computer vision},
  pages={1921--1930},
  year={2019}
}

@inproceedings{son2025maskedkd,
  title={The Role of Masking for Efficient Supervised Knowledge Distillation of Vision Transformers},
  author={Son, Seungwoo and Ryu, Jegwang and Lee, Namhoon and Lee, Jaeho},
  booktitle={European Conference on Computer Vision},
  pages={379--396},
  year={2025},
  organization={Springer}
}

@inproceedings{liu2021swin,
  title={Swin transformer: Hierarchical vision transformer using shifted windows},
  author={Liu, Ze and Lin, Yutong and Cao, Yue and Hu, Han and Wei, Yixuan and Zhang, Zheng and Lin, Stephen and Guo, Baining},
  booktitle={Proceedings of the IEEE/CVF international conference on computer vision},
  pages={10012--10022},
  year={2021}
}

@inproceedings{sun2024logit,
  title={Logit standardization in knowledge distillation},
  author={Sun, Shangquan and Ren, Wenqi and Li, Jingzhi and Wang, Rui and Cao, Xiaochun},
  booktitle={Proceedings of the IEEE/CVF Conference on Computer Vision and Pattern Recognition},
  pages={15731--15740},
  year={2024}
}

@inproceedings{miles2024vkd,
  title={VkD: Improving Knowledge Distillation using Orthogonal Projections},
  author={Miles, Roy and Elezi, Ismail and Deng, Jiankang},
  booktitle={Proceedings of the IEEE/CVF Conference on Computer Vision and Pattern Recognition},
  pages={15720--15730},
  year={2024}
}

@inproceedings{he2016deep,
  title={Deep residual learning for image recognition},
  author={He, Kaiming and Zhang, Xiangyu and Ren, Shaoqing and Sun, Jian},
  booktitle={Proceedings of the IEEE conference on computer vision and pattern recognition},
  pages={770--778},
  year={2016}
}

\setcounter{secnumdepth}{0} 

\newpage

\section{Appendix}

\section{A. Generality of the U-Shaped Entropy Profile in Vision Transformers}

In our main paper, we identify a distinctive ``U-shaped" information processing signature in the CaiT-S24 teacher model, as revealed by its layer-wise Shannon entropy profile. This signature consists of an initial compression phase, where entropy decreases, followed by an expansion phase, where entropy increases. To substantiate our claim that this U-shaped profile is a fundamental and widespread characteristic of ViT architectures, rather than an artifact of a specific model or training approach, this appendix presents additional entropy analyses on other ViT models.

We investigate two additional models:
\begin{enumerate}
	\item A standard ViT model trained using a conventional supervised learning paradigm on the ImageNet-1k dataset.
	\item A ViT model trained using a self-supervised learning approach, specifically the Masked Autoencoder (MAE) method.
\end{enumerate}

The Shannon entropy for each layer was computed using the same methodology described in Section 2.2 of the main paper. Our findings, detailed below, confirm that the U-shaped entropy structure is a consistent operational signature across these diverse models.

\begin{figure}[t]
	\centering
	\includegraphics[width=0.45\textwidth]{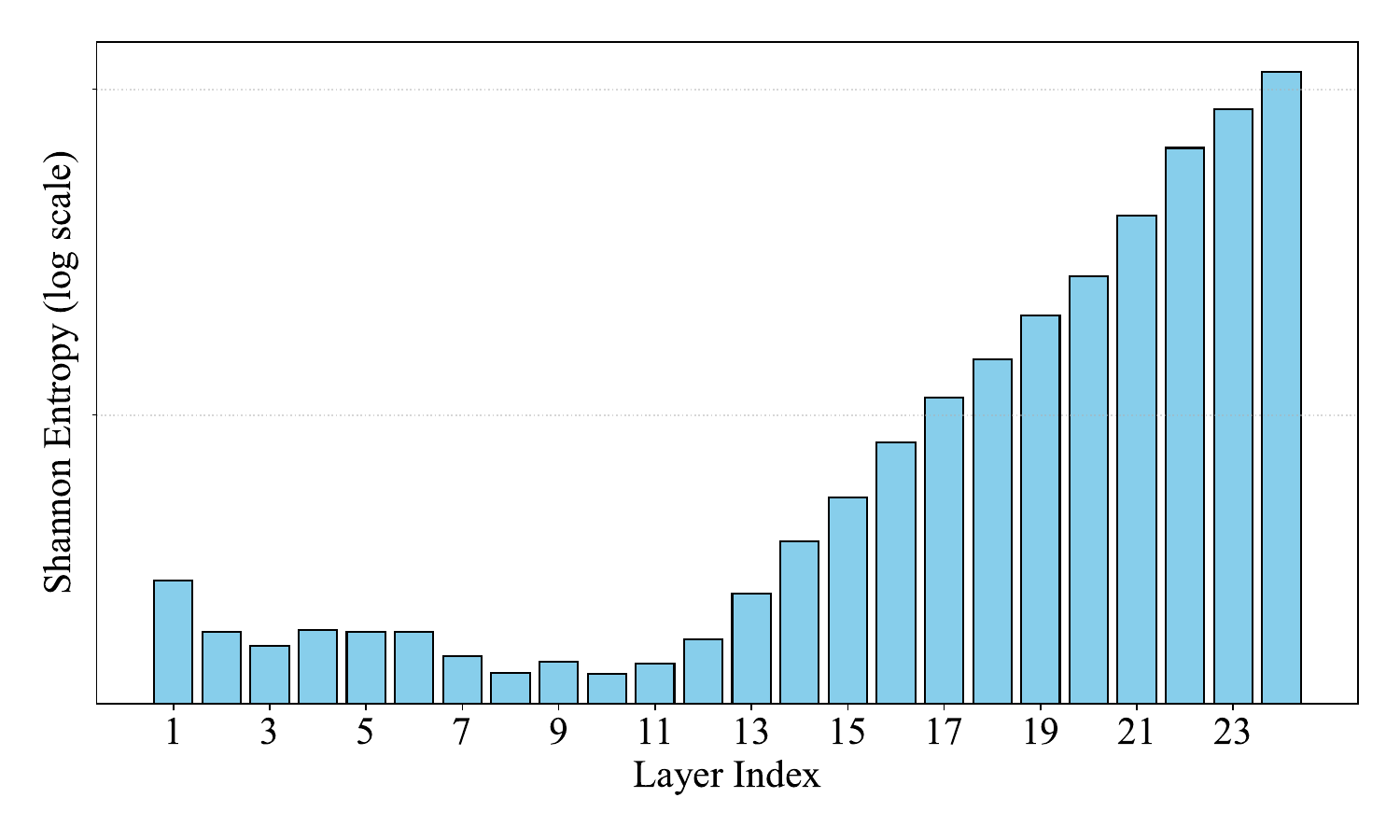} 
	\caption{Layer-wise Shannon entropy for a standard ViT trained with supervised learning. The model displays the characteristic U-shaped profile, indicating an initial compression phase followed by an expansion phase.}
	\label{fig:vit}
\end{figure}

\begin{figure}[t]
	\centering
	\includegraphics[width=0.45\textwidth]{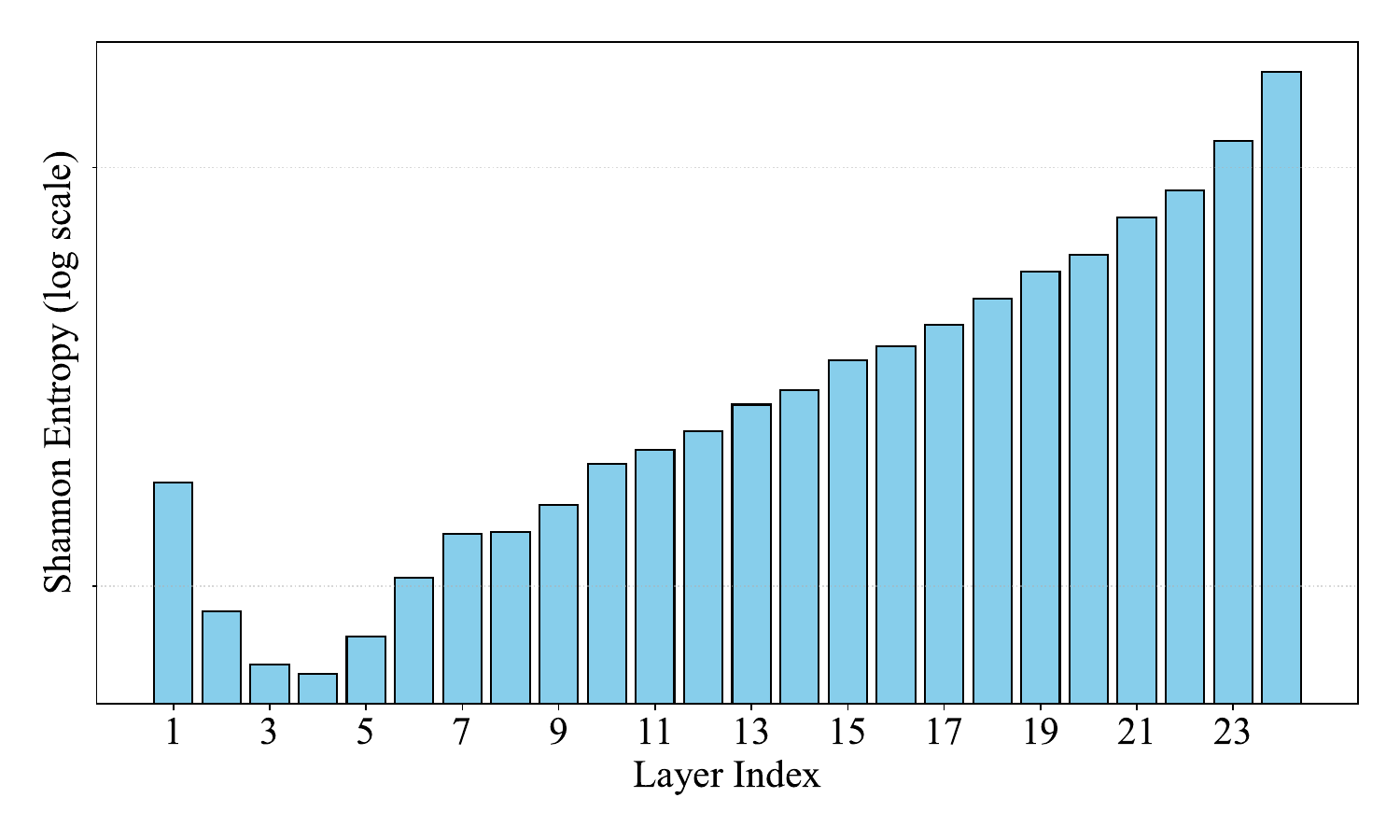} 
	\caption{Layer-wise Shannon entropy for a ViT trained via the Masked Autoencoder (MAE) self-supervised method. This model also exhibits a U-shaped information processing signature, confirming the pattern's independence from the training paradigm. }
	\label{fig:mae}
\end{figure}

\subsection{A.1 U-Shaped Profile in Supervised ViT}

We analyze the layer-wise Shannon entropy of a standard ViT model trained with supervised learning. As illustrated in Figure~\ref{fig:vit}, this model exhibits a distinct U-shaped entropy curve that is highly consistent with the profile of the CaiT-S24 teacher model discussed in the main paper (see Figure 1 in the main paper). The entropy initially decreases through the early and mid-layers, reaching a minimum around layer 9, before steadily increasing through the final layers. This demonstrates that the two-phase processing strategy—compression followed by expansion—is not unique to the CaiT architecture but is also a learned characteristic of vanilla ViTs under standard supervision.

\subsection{A.2 U-Shaped Profile in Self-Supervised ViT (MAE)}

To further test the generality of this information processing strategy, we analyze a ViT trained with the self-supervised MAE methodology. Self-supervised learning provides a different objective function and learning dynamic compared to traditional supervision. Despite these differences, the resulting model's entropy profile, shown in Figure~\ref{fig:mae}, once again displays a clear U-shaped structure. The entropy reaches its nadir around layer 4 and subsequently increases toward the output. While the minimum point of the ``U" is earlier compared to the supervised models, the fundamental two-phase pattern of compression and expansion remains intact.

The additional results presented in this appendix strongly support the central claim that the U-shaped entropy profile is a learned, operational signature inherent to ViT models. This pattern consistently emerges regardless of specific architectural variants (CaiT vs. standard ViT) and training paradigms (supervised vs. self-supervised). The persistence of this compression-then-expansion strategy highlights its role as a fundamental aspect of how ViTs process information. This insight reinforces the main paper's conclusion that a deep understanding of this two-phase dynamic is critical for developing principled and effective knowledge distillation methods for the broader family of Vision Transformer architectures.

\begin{figure*}[t]
	\centering
	\begin{minipage}[b]{\linewidth}
		\subfigure[Swin Transformer First Stage.]{
			\includegraphics[width=0.465\linewidth]{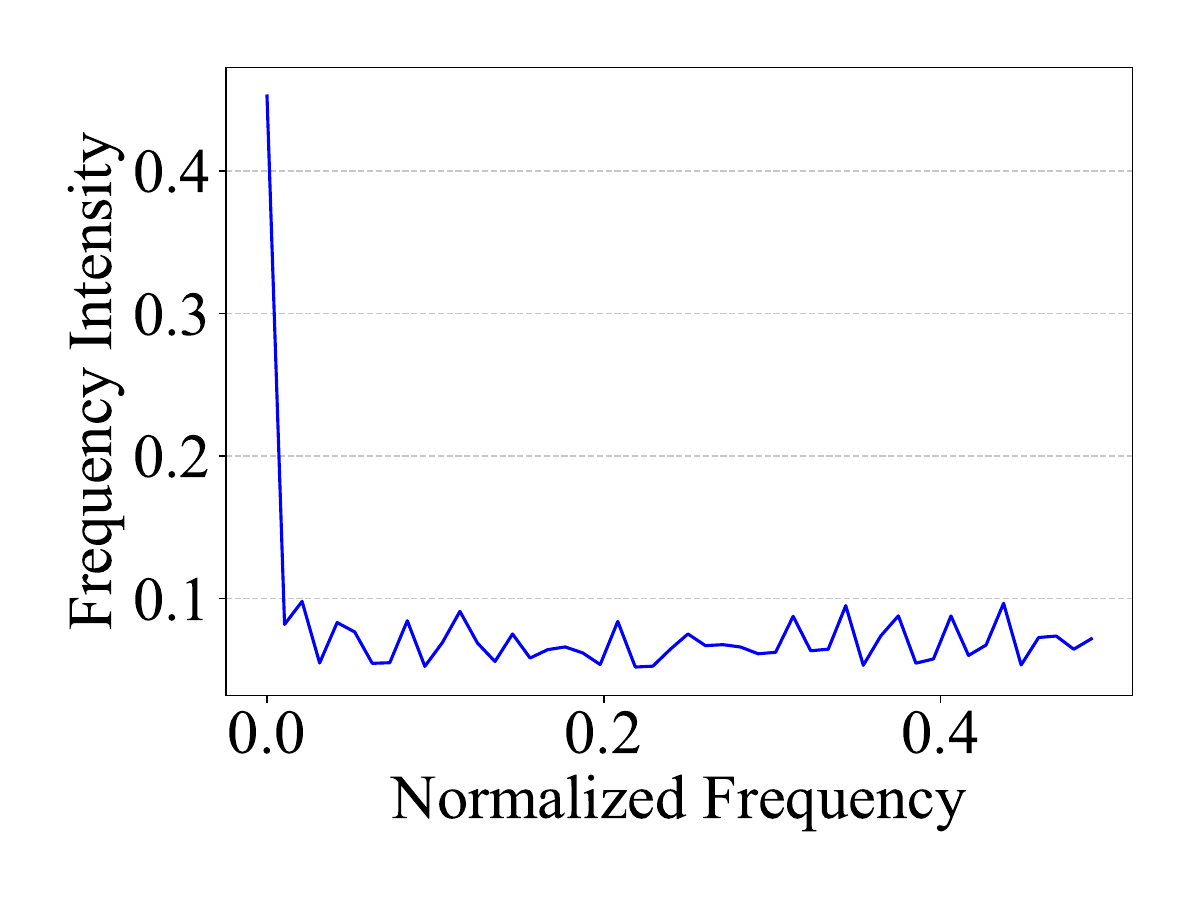}
			\label{fig:spectral_analysis_a}
		}
		\subfigure[Swin Transformer Last Stage.]{
			\includegraphics[width=0.465\linewidth]{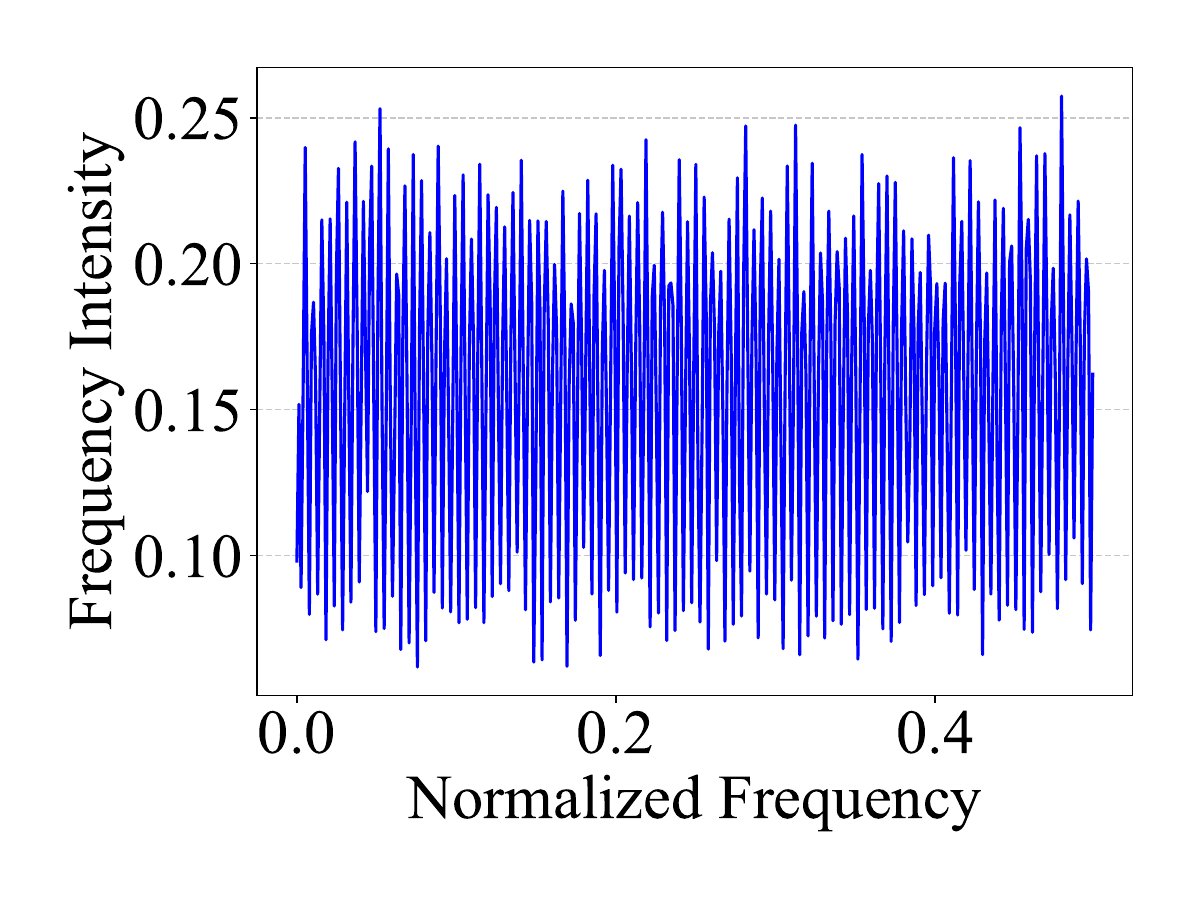}
			\label{fig:spectral_analysis_b}
		}
	\end{minipage}
	\begin{minipage}[b]{\linewidth}
		\subfigure[ResNet101 First Stage.]{
			\includegraphics[width=0.465\linewidth]{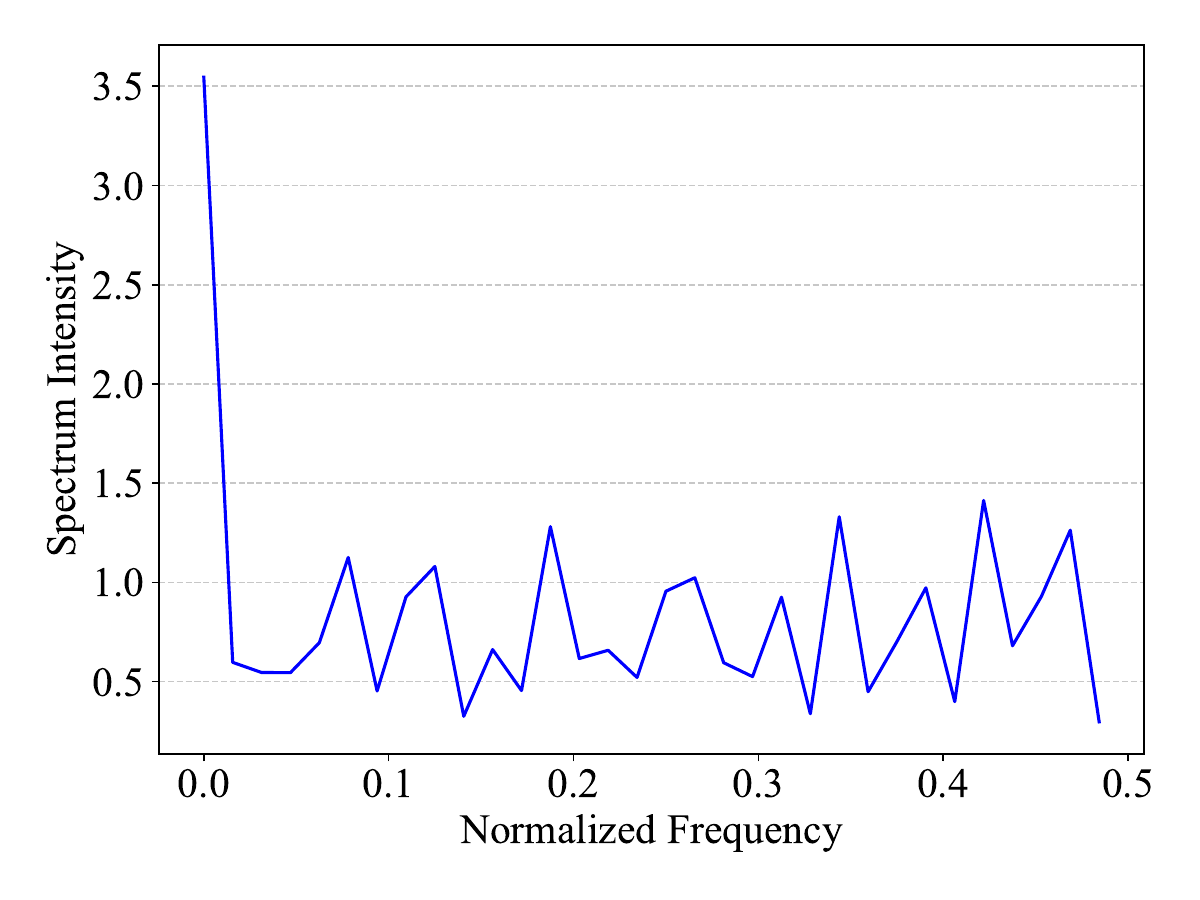}
			\label{fig:spectral_analysis_c}
		}
		\subfigure[ResNet101 Last Stage.]{
			\includegraphics[width=0.465\linewidth]{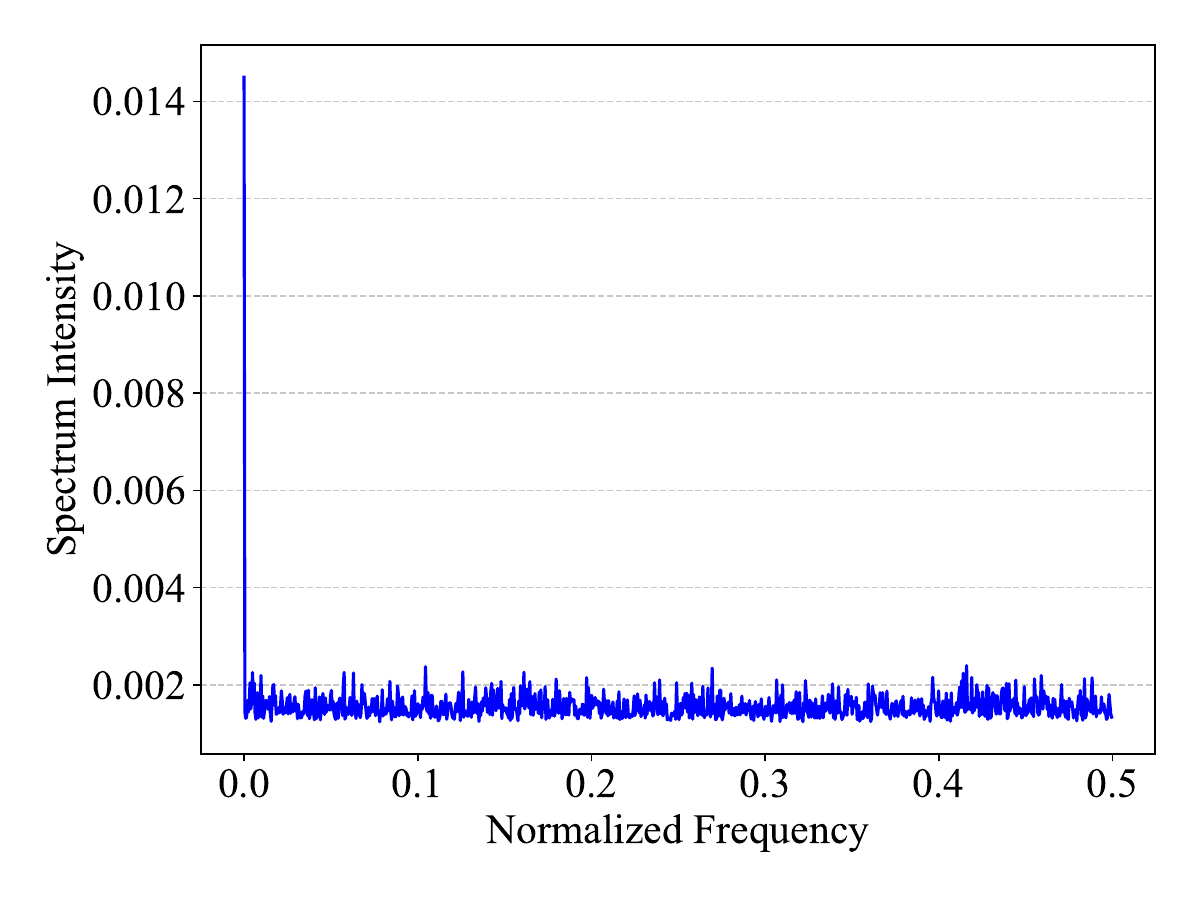}
			\label{fig:spectral_analysis_d}
		}
	\end{minipage}
	\caption{Encoding pattern of First and Last stages in Swin Transformer and ResNet101. (a)-(b) The Swin Transformer evolves from a low-pass spectrum in the first stage to a uniform, high-energy spectrum in the last stage, suggesting full utilization of the channel dimension. (c)-(d) In contrast, ResNet101 maintains a low-pass spectral characteristic in both its first and final stages, indicating its channel capacity is not fully utilized. This fundamental difference in late-layer representation explains why feature mimicry is effective for CNNs but causes negative transfer in ViTs.}
	\label{fig:spectral_analysis}
\end{figure*}

\section{B. Comparative Spectral Analysis: Swin Transformer vs. ResNet101}

To further contextualize our findings on the unique information processing dynamics of Vision Transformers (ViTs), this section provides a comparative spectral analysis of a Swin Transformer and a ResNet101, a representative Convolutional Neural Network (CNN). This analysis utilizes the frequency spectrum method detailed in Section 2.1, which unconventionally examines the feature representations along the channel dimension. The goal is to demonstrate that the representational mismatch we identify is a fundamental difference between the ViT and CNN architectural paradigms.

As shown in the Figure~\ref{fig:spectral_analysis}, the Swin Transformer and ResNet101 exhibit similar overall architectural structures but possess entirely different encoding features in their final layers. The first stage of the Swin Transformer (Figure~\ref{fig:spectral_analysis_a}) shows a spectrum with low-pass characteristics, where energy is concentrated in the lower frequencies. This aligns with the initial compression phase identified in the CaiT model. In its final stage (Figure~\ref{fig:spectral_analysis_b}), however, the spectrum transforms into a uniform, high-energy distribution. This pattern suggests that information is encoded evenly across the spectral space, indicating that the full capacity of the channel dimension  is being utilized. This evolution confirms that transitioning to a distributed, high-dimensional encoding strategy in late layers is a consistent characteristic of ViT architectures.

In contrast, the ResNet101 model shows a different spectral evolution. While its first stage also displays a low-pass pattern (Figure~\ref{fig:spectral_analysis_c}), its final stage (Figure~\ref{fig:spectral_analysis_d}) retains this low-pass characteristic. The spectral energy remains concentrated in low-frequency components, with a significantly attenuated overall signal intensity. This indicates that even in its deepest layers, the CNN representation remains compact and does not expand to exploit the full dimensionality of its channel space.

\subsection{B.1 Implications for Knowledge Distillation}

This direct comparison reveals a crucial divergence in representational strategy. While ViTs transition from a compressed to a distributed encoding paradigm, CNNs maintain a compact, low-pass encoding throughout their depth. This underutilization of the channel dimension in CNNs explains the differing efficacy of feature-based distillation across architectures.

A smaller student CNN can successfully mimic a teacher's final-layer features because they share the same compact representational paradigm. For ViTs, however, forcing a student model with limited channel capacity to mimic the distributed, high-dimensional encoding of a large teacher creates a fundamental ``representational paradigm mismatch". This mismatch provides a conflicting supervisory signal that is architecturally incompatible with the student, causing the negative transfer we observe in our experiments. The success of feature distillation in CNNs and its failure in ViTs is therefore not an empirical result but a direct consequence of their differing late-layer encoding strategies.

\section{C. Distillation Methods for Validation}

To validate the insights derived from our analytical framework, we introduce and evaluate two distinct feature-based distillation methods. These methods are compared against a standard logits-only baseline to assess their efficacy. The first, which we term SpectralKD, is a non-parametric approach that explicitly aligns the frequency spectra of the student and teacher models. The second, ProjectorKD, is a parametric method that uses a learnable projector to match feature maps, drawing inspiration from the seminal FitNet architecture.

\subsection{C.1 Frequency Alignment (SpectralKD)}

Motivated by our analytical findings, SpectralKD explicitly aligns the spectral properties of the student and teacher models. This method operates on the principle that matching representations in the frequency domain can enforce a more profound structural similarity than simple point-wise feature imitation. It can be conceptualized as a form of relational or structural distillation, as it constrains the student to replicate the spatial-frequency relationships within the teacher's feature maps.

The core of this method is the application of a 2D Fast Fourier Transform (FFT) to the spatial dimensions ($H \times W$) of the intermediate feature maps, denoted as $\mathbf{A}_s$ and $\mathbf{A}_t$ for the student and teacher respectively. This choice directly targets the spatial interplay between tokens, which is central to the self-attention mechanism's function. By matching in this domain, the method enforces alignment of both low-frequency components (representing global, coarse-grained structures) and high-frequency components (corresponding to local, fine-grained details). The implementation follows three steps:

\begin{enumerate}
	\item \textbf{Channel Alignment.} If the student and teacher feature maps,  $\mathbf{A}_s$ and $\mathbf{A}_t$, have different channel dimensions ($C_s \neq C_t$), we use adaptive average pooling to downsample the feature map with the larger channel dimension to match the smaller one, resulting in a shared channel dimension $C = \min(C_s, C_t)$. This is a simple, parameter-free technique to resolve dimensional mismatches while preserving spatial information.
	\item \textbf{Frequency Transform.} We apply the 2D real FFT (RFFT) to the spatial dimensions of the channel-aligned feature maps. The resulting complex tensors are then separated into their real and imaginary parts and stacked to form a real-valued tensor representation $\mathcal{F}_{\mathrm{stack}}(\mathbf{A})$.
	\item \textbf{Loss Calculation.} The frequency alignment loss $\mathcal{L}_{\mathrm{Freq}}$ is calculated as the Mean Squared Error (MSE) between the student's and teacher's frequency representations:
	\begin{equation}
		\mathcal{L}_{\mathrm{Freq}} = \mathrm{MSE}(\mathcal{F}_{\mathrm{stack}}(\mathbf{A}_s), \mathcal{F}_{\mathrm{stack}}(\mathbf{A}_t))
	\end{equation}
\end{enumerate}

The total training objective combines this structural loss with the standard knowledge distillation (KD) loss on the logits. The standard KD loss is defined as:

\begin{equation}
	\begin{aligned}
		\mathcal{L}_{\text{KD}} &= (1-\alpha) \mathcal{L}_{\text{CE}}(f_{s}(\mathbf{x}), y) \\
		&+ \alpha T^{2} \mathcal{L}_{\text{KL}}\left(\frac{f_{s}(\mathbf{x})}{T}, \frac{f_{t}(\mathbf{x})}{T}\right),
	\end{aligned}
\end{equation}
where $\mathcal{L}_{\text{CE}}$ is the cross-entropy loss between the student predictions $f_{s}(\mathbf{x})$ and ground-truth labels $y$, and $\mathcal{L}_{\text{KL}}$ is the Kullback-Leibler divergence. The temperature $T$ smooths the logits of teacher, and $\alpha$ balances the two terms.

The final objective for SpectralKD is: 

\begin{equation} \label{eq:loss}
	\mathcal{L}_{\mathrm{Total}} = \mathcal{L}_{\mathrm{KD}} + \beta \mathcal{L}_{\mathrm{Freq}},
\end{equation}
where $\beta$ is a weighting coefficient. This objective encourages the student to mimic not only the teacher's output distribution but also its internal spatial feature structure.

\begin{table*}[t]
	\centering
	\begin{tabular}{lllc}
		\hline
		\textbf{Method} & \textbf{Teacher Model} & \textbf{Student Model} & \textbf{Top-1 Accuracy (\%)} \\
		\hline
		Baseline (No Distillation) & N/A & DeiT-Tiny & 74.86 \\
		Logits Distillation Only & ViT-Base (MAE, no ft) & DeiT-Tiny & 72.35 \\
		SpectralKD (Logits + Last Layer) & ViT-Base (MAE, no ft) & DeiT-Tiny & 67.05 \\
		Logits Distillation Only & ViT-Large & DeiT-Tiny  &76.85 \\
		SpectralKD (Logits + Last Layer) & ViT-Large & DeiT-Tiny & 76.58 \\
		\hline
	\end{tabular}
	\caption{ImageNet Top-1 accuracy of DeiT-Tiny under different distillation settings.}
	\label{table:mae_validation}
\end{table*}

\subsection{C.2 Projector-Based Feature Distillation (ProjectorKD)}

To provide a comparative baseline, we also implement a parametric feature distillation method inspired by FitNet. This approach, ProjectorKD, serves as a counterpoint to our non-parametric frequency alignment. Instead of operating in the frequency domain, this method functions directly on spatial feature maps.

A learnable projector is added after the student's intermediate layer. This projector transforms the student's feature map $\mathbf{A}_s$ into a new map  $\mathbf{A}^{\prime}_s$ that matches the dimensional shape of the teacher's feature map $\mathbf{A}_t$. The distillation loss is then computed as the MSE between the teacher's features and the student's projected features:

\begin{align}
	\mathbf{A}_{s}^{\prime} & = \mathrm{Projector} (\mathbf{A}_{s})\\
	\mathcal{L}_{\mathrm{Proj}} & = \mathrm{MSE} (\mathbf{A}_{s}^{\prime}, \mathbf{A}_{t})
\end{align}

Similar to SpectralKD, the final loss combines this with the standard KD loss:

\begin{equation} \label{eq:loss}
	\mathcal{L}_{\mathrm{Total}} = \mathcal{L}_{\mathrm{KD}} + \beta \mathcal{L}_{\mathrm{Proj}},
\end{equation}

Unlike the explicit structural constraint imposed by frequency alignment, this parametric method allows the model to learn the optimal linear transformation for mimicking the teacher's representations.

\subsection{C.3 Experimental Setup}

\textbf{Dataset and Models.} We conduct experiments on the ImageNet-1k dataset, which comprises 1.28M training images and 50K validation images across 1,000 classes.

\textbf{Implementation Details.} All experiments are conducted on $4$ NVIDIA RTX 4090D GPUs with a batch size of $512$. Our implementation is based in PyTorch, and we build upon the \emph{timm} library for model architectures and pretrained weights.

\textbf{Hyperparameters.} Unless otherwise stated, we use a distillation temperature of $T=1$,  a knowledge distillation weight of $\alpha  = 0.9$, and a feature distillation weight of $\beta = 0.2$ for all experiments.

\section{D. Validation with More Teachers}

To further validate our claim that feature-based distillation in ViTs can induce negative transfer, we conduct an additional study using two teachers for a DeiT-Tiny student, as summarized in Table~\ref{table:mae_validation}\footnote{The baseline result in Table~\ref{table:mae_validation} was recomputed with our own training setup; it thus differs numerically from the value in the first arXiv version.}. Following our main setup, we first consider a ViT-Base model that is pre-trained with MAE but \emph{not} fine-tuned on ImageNet. In this case, the teacher's logits are essentially uninformative for the downstream classification task. Consistently, distilling from these noisy logits already hurts the student, reducing Top-1 accuracy from $74.86\%$ (no distillation) to $72.35\%$. When we further impose last-layer feature alignment via SpectralKD, performance drops dramatically to 67.05\%. This shows that forcing the student to mimic the late-layer feature maps of an uninformative teacher amplifies negative transfer rather than providing a useful signal.

We then use a supervised ViT-Large teacher. Here, logits-based distillation substantially improves the DeiT-Tiny student, achieving $76.85\%$ Top-1 accuracy. However, adding last-layer feature-map distillation again slightly degrades performance ($76.58\%$ with SpectralKD), even though the teacher itself is highly accurate. Taken together, these results reinforce our central conclusion: late-layer feature-map distillation in ViTs tends to be detrimental, regardless of whether the teacher is weak (MAE, no fine-tuning) or strong (ViT-Large). In both cases, enforcing similarity of high-dimensional feature maps pushes the student toward a representational regime it cannot realize, leading to negative transfer.

\end{document}